\documentclass[conf]{IEEEtran}
\usepackage{algorithm,algpseudocode}
\usepackage{amsmath}
\usepackage{amssymb}
\usepackage{mathtools}
\usepackage[dvipsnames]{xcolor}
\usepackage{tikz}

\newcommand*\circled[1]{\tikz[baseline=(char.base)]{
            \node[shape=circle,draw,inner sep=2pt] (char) {#1};}}
\usepackage{pifont}
\usepackage{cite}
\def\BibTeX{{\rm B\kern-.05em{\sc i\kern-.025em b}\kern-.08em
    T\kern-.1667em\lower.7ex\hbox{E}\kern-.125emX}}
\usepackage[bookmarksnumbered=true]{hyperref} 
\hypersetup{
     colorlinks = true,
     linkcolor = blue,
     anchorcolor = blue,
     citecolor = blue,
     filecolor = blue,
     urlcolor = blue
     }

\usepackage{graphicx}
\usepackage{svg}

\usepackage{wrapfig}
\usepackage{caption}
\usepackage{subcaption}
\usepackage{booktabs}

\usepackage{tikz}

\usepackage{ragged2e}
\usepackage{siunitx}
\usepackage{multicol}
\usepackage{multirow}

\usepackage{xcolor}
\begin{document}
\title{TReX- \underline{Re}using Vision \underline{T}ransformer's Attention for Efficient \underline{X}bar-based Computing} 

\author{Abhishek Moitra, Abhiroop Bhattacharjee, Youngeun Kim, and Priyadarshini Panda\\Department of Electrical Engineering, Yale University, New Haven, CT 06511} 

\maketitle 


\begin{abstract}
Due to the high computation overhead of Vision Transformers (ViTs), In-memory Computing architectures are being researched towards energy-efficient deployment in edge-computing scenarios. 
Prior works have proposed efficient algorithm-hardware co-design and IMC-architectural improvements to improve the energy-efficiency of IMC-implemented ViTs. However, all prior works have neglected the overhead and co-depencence of attention blocks on the accuracy-energy-delay-area of IMC-implemented ViTs. To this end, we propose TReX- an attention-reuse-driven ViT optimization framework that effectively performs attention reuse in ViT models to achieve optimal accuracy-energy-delay-area tradeoffs. TReX optimally chooses the transformer encoders for attention reuse to achieve near-iso-accuracy performance while meeting the user-specified delay requirement. Based on our analysis on the Imagenet-1k dataset, we find that TReX achieves 2.3$\times$ (2.19$\times$) EDAP reduction and 1.86$\times$ (1.79$\times$) TOPS/mm$^2$ improvement with $\sim$1\% accuracy drop in case of DeiT-S (LV-ViT-S) ViT models. Additionally, TReX achieves high accuracy at high EDAP reduction compared to state-of-the-art token pruning and weight sharing approaches. On NLP tasks such as CoLA, {TReX leads to 2\% higher non-ideal accuracy compared to baseline at 1.6$\times$ lower EDAP.}

\end{abstract}

\begin{IEEEkeywords}
Vision Transformers, Attention sharing, In-memory Computing
\end{IEEEkeywords}

\IEEEpeerreviewmaketitle
\section{Introduction}
\begin{figure}[t]
    \centering
    \includegraphics[width=\linewidth]{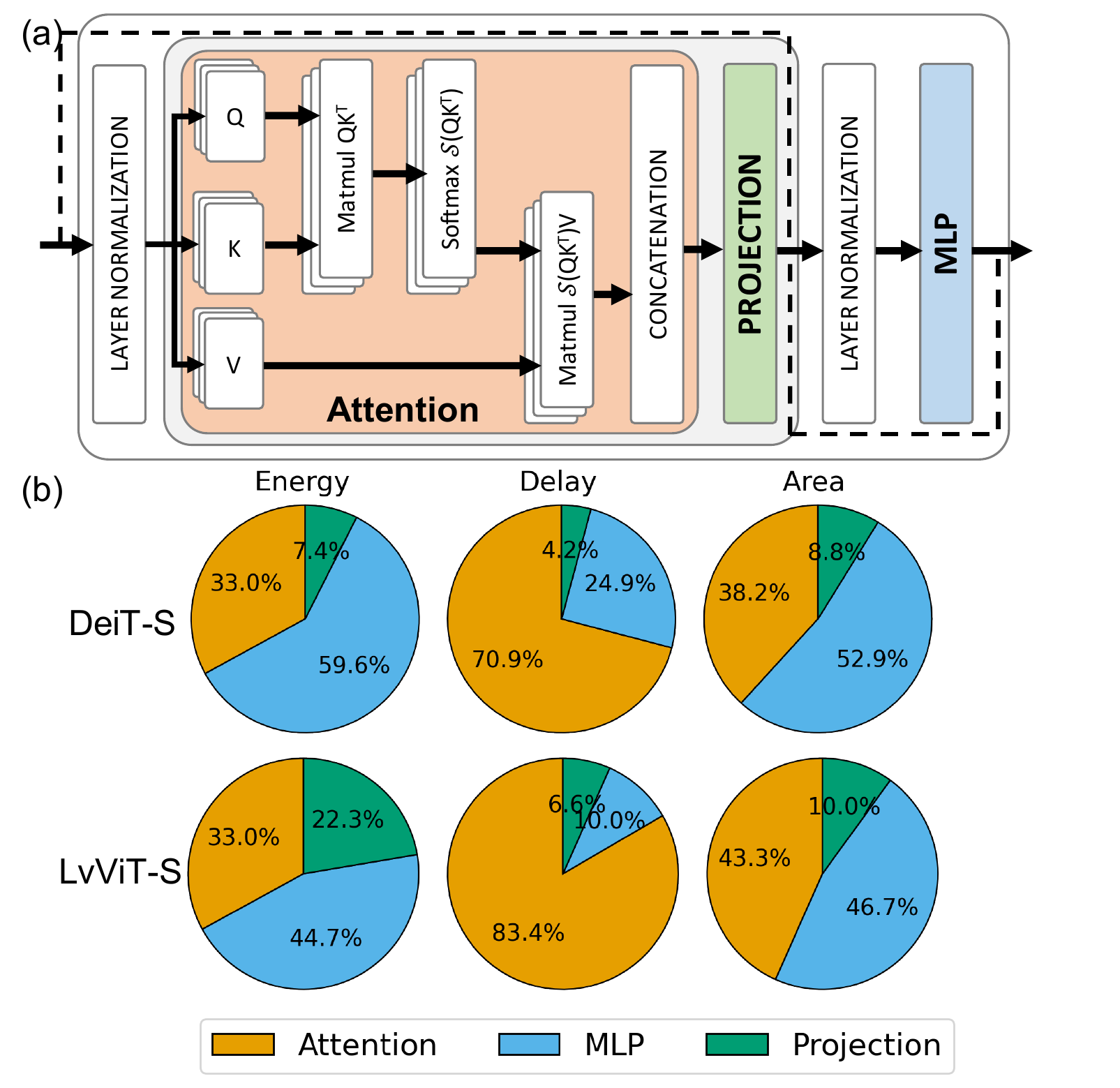}
    \caption{(a) Encoder architecture of a vision-transformer \cite{jiang2021all,touvron2021training}. (b) Pie diagrams of energy, delay and area for DeiT-S \cite{touvron2021training} and LV-ViT-S \cite{jiang2021all} transformer encoders. The energy, delay and area correspond to 64$\times$64 FeFET crossbar implementations. }
    \label{fig:EDA_Pie_FeFET}
\end{figure}

The high accuracy and robustness of vision transformers (ViT) \cite{touvron2021training, zhai2022scaling, ranftl2021vision} on large-scale image recognition tasks have made them suitable candidates for edge intelligence \cite{maaz2023edgenext, chen2023efficient}. However, as seen in Fig. \ref{fig:EDA_Pie_FeFET}a, the huge computation overhead of ViTs due to feed-forward operations (layers $Q$, $K$, $V$, \textit{Projection}, \textit{MLP}), softmax operation ($\mathcal{S}(QK^T)$) and matrix multiplications ($Matmul~QK^T$ and $Matmul~\mathcal{S}(QK^T)V$, where $\mathcal{S}$ denotes softmax) pose a major roadblock towards resource-constrained edge-deployment. To this end, In-Memory Computing (IMC) architectures have been proposed as an energy-efficient alternative to traditional von-Neumann computing architectures for edge-computing. {Compared to von-Neumann architectures, IMC effectively mitigates weight-specific data movement between memories, thereby alleviating the memory-wall bottleneck \cite{moitra2023xpert, chen2018neurosim}. Additionally, IMC architectures enable highly parallel multiply-accumulate computations per cycle in an energy and area-efficient manner, leveraging the analog nature of computing \cite{chen2018neurosim, moitra2023xpert}.} 
\begin{figure}[h!]
    \centering
    \includegraphics[width=\linewidth]{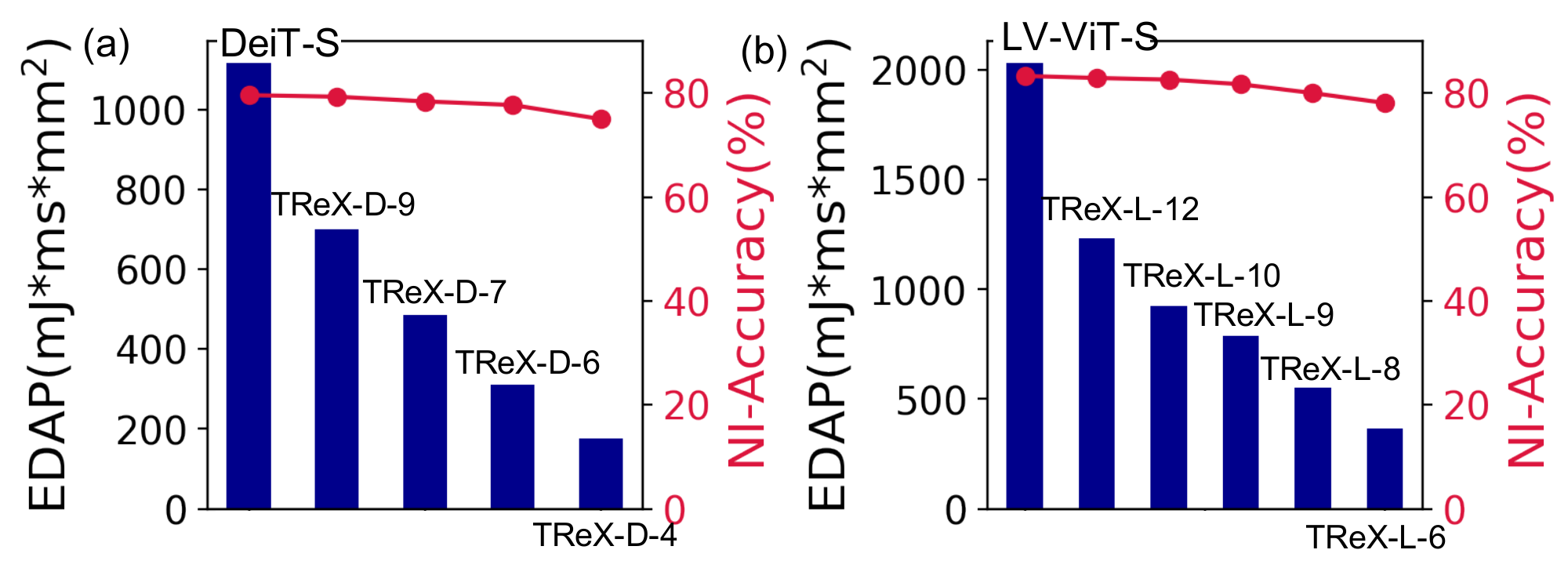}
    \caption{Figure showing the EDAP and non-ideal accuracy of (a) Baseline DeiT-S \cite{rao2021dynamicvit} and TReX-D-$N$ ViTs. (b) Baseline LV-ViT-S \cite{rao2021dynamicvit} and TReX-L-$N$ ViTs. Here, TReX-D-$N$ and TReX-L-$N$ denote TReX-optimized DeiT-S and LV-ViT-S ViTs, respectively optimized at different target delay values $N$. Non-ideal accuracy denotes the hardware accuracy measured on FeFET-based IMC implementation with ADC/device variation noise. Accuracy measured for ImageNet-1K dataset.}
    \label{fig:prelim_results}
\end{figure}

However, as seen in Fig. \ref{fig:EDA_Pie_FeFET}b, the attention module in IMC-implemented ViTs consume high energy, delay and area overhead due to fully connected, $Matmul$ and softmax layers \cite{yang2022full, yang2020retransformer}. The energy consumption of the attention block is attributed to the crossbar read operations over a large number of tokens and the softmax ($\mathcal{S}(QK^T)$) operations (See Section \ref{sec:Trex_sim} for more details). Similarly, the area is attributed to the large number of crossbars required by the $Q$, $K$, $V$, and the $Matmul$ layers (See Section \ref{sec:Trex_sim}). Interestingly, IMC-based implementation of $QK^T$ and $\mathcal{S}(QK^T)V$ layers, require crossbar write operations \cite{yang2020retransformer, yang2022full}. The write operations inside the attention block makes it a significant contributor ($\sim$80\%) to the overall delay. 
Thus, there is a need for IMC-aware ViT co-design to achieve low power, high throughput and area-efficient edge deployment.

Over the recent years, there have been several algorithmic \cite{rao2021dynamicvit, dong2023heatvit, zheng2022savit,zhang2022minivit,lin2022spin,chen2021psvit} and IMC-centric \cite{zhang2023xformer, yang2022full, yang2020retransformer} approaches to ameliorate the computation overhead of ViTs. Algorithmic approaches like token pruning \cite{rao2021dynamicvit, dong2023heatvit} is an optimization strategy where redundant input tokens are pruned out in order to achieve energy-efficiency at iso-accuracy with that of an unpruned model. Other works have proposed head pruning \cite{zheng2022savit} wherein, the attention heads of the multi-head self-attention (MHSA) blocks are pruned out in order to minimize the number of operations and maintain iso-accuracy. Weight sharing \cite{zhang2022minivit, lin2022spin} has also emerged as a standard transformer optimization strategy wherein multiple encoders in a transformer share the same weight parameters. {Recent IMC-centric transformer co-optimization works \cite{yang2020retransformer, yang2022full, sridharan2023x} have proposed IMC architecture implementations to efficiently compute the $Matmul$ layers. Additionally, works like X-Former \cite{zhang2023xformer} have proposed sequence blocking dataflow to increase the parallelism of transformer layers. }Furthermore, the authors in a recent work \cite{spoon2021toward}, have proposed variation-aware training of transformers to achieve higher robustness against IMC quantization and device variation noise. 


However, prior algorithmic and IMC-centric co-design approaches have several drawbacks. Algorithmic co-optimization works tend to achieve high accuracy only at low pruning and weight sharing ratios (with small hardware efficiency). Similarly, IMC-centric works such as \cite{yang2020retransformer, yang2022full} neglect the crossbar write energy and delay during $Matmul$ operations thereby achieving sub-optimal results. Finally, all prior works have neglected the co-dependence of attention blocks on the accuracy-energy-delay-area of IMC-implemented ViTs. 





To this end, we propose TReX- a framework to perform effective attention reuse in ViTs to achieve good accuracy-energy-delay-area tradeoff. In the attention reuse technique, the attention block's output of one encoder (\textit{i.e.,} output of the \textit{concatenation} operation) is reused in the attention block of the subsequent encoder (Fig. \ref{fig:attention_sharing_example}). For attention reuse, we use a small transformation block (TB) that introduces data variability (thus, maintaining good accuracy) while adding minimal hardware overhead. TReX optimally chooses the transformer encoders for attention reuse to achieve near-iso-accuracy performance while meeting the user-specified delay requirement. As shown in Fig. \ref{fig:prelim_results}, {TReX achieves near-iso-accuracy while significantly reducing the energy-delay-area product (EDAP) for DeiT-S and LV-ViT-S transformer models. }
The key contributions of our work are:
\begin{enumerate}
    \item We propose TReX- an optimization framework that effectively performs attention reuse in ViT models to achieve optimal accuracy-energy-delay-area tradeoffs.
    
    \item We propose TReXSim- an IMC-realistic benchmarking platform that can be used to evaluate different ViT models implemented with different IMC devices (such as FeFET, SRAMs). The TReX framework uses TReXSim to optimally choose the number of encoders for attention reuse based on a user-defined delay requirement. 

    \item We perform our analysis on state-of-the-art compact ViT models such as DeiT-S \cite{rao2021dynamicvit} and LV-ViT-S \cite{yang2022full}. Based on our analysis on the Imagenet-1k dataset, we find that TReX achieves 2.3$\times$ (2.19$\times$) EDAP reduction and 1.86$\times$ (1.79$\times$) TOPS/mm$^2$ improvement with $\sim$1\% accuracy drop in case of DeiT-S (LV-ViT-S) ViT models. Additionally, we find that TReX can effectively achieve high accuracy at high EDAP reduction compared to state-of-the-art token pruning \cite{rao2021dynamicvit} and weight sharing \cite{zhang2022minivit} approaches.

    
\end{enumerate}

\section{Related Work}
\textbf{ViT Optimization Works:} Works such as \cite{rao2021dynamicvit, dong2023heatvit, xu2022evo, meng2022adavit } have proposed token pruning wherein redundant tokens have been pruned out to improve the efficiency. Some recent works have  adopted token merging techniques wherein redundant tokens are systemically merged into a single token to achieve accuracy-efficiency tradeoff \cite{dong2023heatvit}. Both token pruning approaches use additional predictor networks to distinguish the important and non-informative tokens to achieve low latency while maintaining good accuracy. Recent works \cite{zheng2022savit, shen2022lottery} have performed static pruning of weights and attention head combined with token pruning to achieve higher efficiency. Additionally, weight sharing approaches \cite{zhang2022minivit, lin2022spin} have been proposed to reduce the number of parameters required by ViTs. Here, the weights of one encoder are shared between multiple encoders. While weight sharing compresses the ViT model, it does not reduce the computation overhead. There have been several works \cite{wang2022energy, wang2021spatten} that have proposed sparsifying the attention blocks by pruning the heads and the tokens. However, these works only partially overcome the attention overhead and require sparse computation hardware to efficiently implement the sparse attention operation.

TReX is a ViT optimization strategy that differs from prior optimization works in several aspects. While prior works focus on the token and parameter overhead of ViTs, TReX targets the compute and area expensive attention blocks. While ViT token pruning methods require several predictor models to select important and eliminate non-informative tokens, TReX uses attention reuse in ViTs that does not require any extra computation overhead. Unlike prior weight sharing techniques that only reduce the parameter overhead of ViTs and do not reduce the computation overhead, TReX effectively ameliorates both computation and parameter overhead of ViTs.

\textbf{IMC-implementations of Transformers:} Recently, many works have proposed efficient IMC implementations for transformers \cite{yang2022full, yang2020retransformer}. The authors in \cite{yang2022full} propose fully analog implementations for transformers by using memristive circuits for dot-product operations and analog circuits for implementing non-linear functions such as GeLU, ReLU and softmax. Another work ReTransformer \cite{yang2020retransformer} proposes matrix decomposition techniques to eliminate the bottlenecks introduced during transformer operation. However, prior IMC-implementations of transformers have neglected the energy and delay overheads of NVM device write operations during the $Matmul$ operations. Considering the write overhead of NVM devices, prior IMC-centric transformer approaches \cite{yang2020retransformer, yang2022full} are incapable of achieving optimal performance.
\section{Background on Vision Transformers}
\label{sec:vision-transformers}

A Vision Transformer (ViT) model proposed in \cite{touvron2021training, jiang2021all} segments an image into multiple patches, commonly referred to as tokens. Depending on the patch size, the number of tokens $t$ is determined. Each token is embedded into a $d$-sized feature space. Thus, the input to an encoder $X$ is a $t\times d$ dimension vector. A ViT consists of multiple encoders determined by the $N_{Encoders}$ parameter. 

Fig. \ref{fig:EDA_Pie_FeFET}a shows the architecture of a ViT encoder. In each encoder, the inputs $X$ of dimensions $t\times d$ are multiplied with the weights $W_Q$, $W_K$ and $W_V$ to generate the Query (Q), Key (K) and Value (V) matrices. ViTs use multi-head self-attention (MHSA)-based encoder, that capture closer relationships between the Query and Key values \cite{touvron2021training, jiang2021all, zhang2022minivit, rao2021dynamicvit}. For this, the Q, K and V outputs are partitioned into smaller singular-heads ($Q_i$, $K_i$, $V_i$), where $i$ denotes a head of MHSA.

The attention shown in Equation \ref{eq:attention_eq} is computed using matrix multiplications between $Q_i$, $K_i^T$ followed by Softmax operation and finally matrix multiplication with $V_i$. In TReX, to compute the softmax of each element $x_i$ in a matrix $x$, we subtract $x_{max}$ (the maximum value in $x$) from $x_i$ as shown in Equation \ref{eq:softmax}. This is numerically more stable compared to the standard softmax function as it eliminates any overflow/underflow issues during digital computation \cite{stevens2021softermax, yang2020retransformer}.

\begin{equation}
Attention(Q_i,K_i,V_i) = {Softmax(\frac{Q_iK_i^T}{\sqrt{d}})}V_i,
\label{eq:attention_eq}
\end{equation}
\begin{equation}
    Softmax(x_i) = \frac{e^{x_i-x_{max}}}{\sum_i {e^{x_i-x_{max}}}}.
    \label{eq:softmax}
\end{equation}

Next, the attention outputs are concatenated resulting in a $t\times d$ output attention matrix. Following this, the projection and MLP layers project the information into a higher dimension feature space. Each encoder outputs a $t\times d$ vector that is forwarded to the subsequent encoder.
For convinience, in the rest of the paper, we will use $QK^T$ to represent $Q_iK_i^T$, $\mathcal{S}(QK^T)$ to represent $\mathcal{S}(Q_iK_i^T)$ and $\mathcal{S}(QK^T)V$ to represent the $Attention(Q_i,K_i,V_i)$ in Equation \ref{eq:attention_eq}. 
\section{In-Memory Computing Architectures}
\begin{figure}
    \centering
    \includegraphics[width=0.8\linewidth]{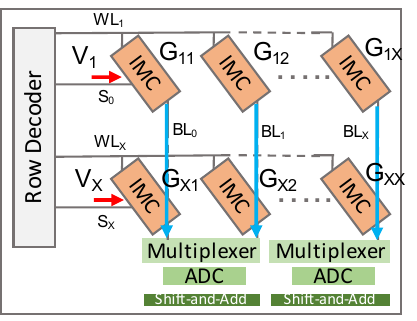}
    \caption{Figure showing a standard IMC crossbar array.}
    \label{fig:xbar_img}
\end{figure}
Analog crossbars \cite{moitra2023xpert, moitra2023spikesim, chen2018neurosim} consist of 2D arrays of In-Memory-Computing (IMC) devices and several peripheral circuits such as row-decoders, multiplexers (for resource sharing), Analog-to-Digital Converters (ADCs), shift-and-add circuits along with write circuits responsible for programming the IMC devices. The neural network's activations are bit-serialized and represented as analog voltages $V_i$ and are fed into each source line $S_i$ of the crossbar, while the IMC devices are programmed to specific conductance values ($G_{ij}$) to represent the network weights, as shown in Fig. \ref{fig:xbar_img}. During inference in an ideal crossbar, the voltages interact with the device conductances, resulting in a current governed by Ohm's Law. These currents are generated along the bit-lines $BL_i$. As per Kirchoff's current law, the net output current sensed at each column $j$ is the summation of currents through each device, expressed as $I_{j(ideal)} = \Sigma_{i=1}^{X}{G_{ij} * V_i}$. Here, $G_{ij}$ is the matrix containing the ideal conductance values of the IMC devices and $X$ is the crossbar size. The analog currents are converted to digital values using the ADC and accumulated using the shift-and-add circuits. 

In practice, due to the analog nature of computation, various hardware noise and non-idealities arise, such as interconnect parasitic resistances, synaptic device-level variations during read and write operations \cite{jain2020rxnn, bhattacharjee2023examining, chakraborty2020geniex} among others. As a result, the non-ideal current $I_{j(non-ideal)}$ is represented as $I_{j(non-ideal)} = \Sigma_{i=1}^{X}{G_{ij}' * V_i}$. The $I_{j(non-ideal)}$ deviates from its ideal value owing to the non-ideal conductance matrix $G_{ij}'$. This leads to significant accuracy losses for neural networks mapped onto crossbars, especially for larger crossbars with more non-idealities \cite{jain2020rxnn,  chakraborty2020geniex, bhattacharjee2022examining}.
\begin{figure*}[h!]
    \centering
    \includegraphics[width=0.9\textwidth]{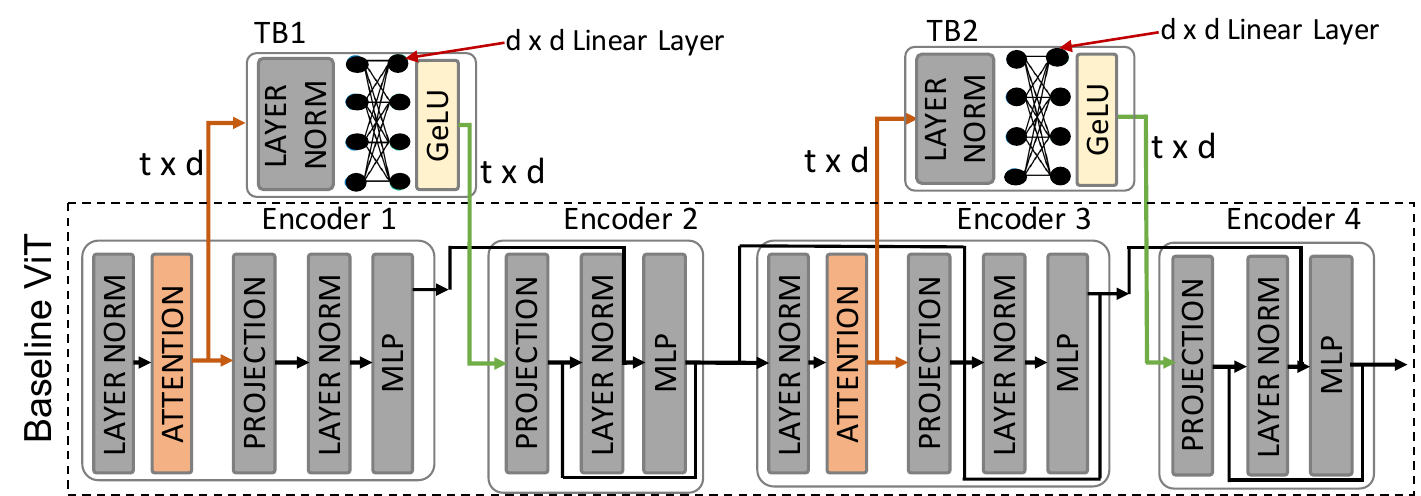}
    \caption{Figure showing the attention reuse technique wherein encoders 2 and 4 reuse attention of encoders 1 and 3, respectively. Transformation blocks TB1 and TB2 are used for introducing data variability. For encoders 2 and 4, the Attention blocks are removed and the TB outputs are connected to the Projection input.}
    \label{fig:attention_sharing_example}
\end{figure*}
\section{TReX Optimization Framework}
\subsection{Attention Reuse Technique}
Fig. \ref{fig:attention_sharing_example} shows an example of attention reuse in a \textit{BaseViT} with 4 encoders (\textit{i.e.,} $N_{Encoders}=4$). Here, encoders 2 and 4 reuse the attention (concatenation output) of encoders 1 and 3, respectively. For attention reuse, transformation blocks TB1 and TB2 are introduced. Transformation blocks contain a layer-normalization layer, a $d\times d$ fully-connected layer and a GeLU layer. TB1 and TB2 transform the $t\times d$ attention outputs of encoders 1 and 3 to a $t\times d$ vector which is fed to the projection blocks of encoders 2 and 4, respectively. Here, $t$ and $d$ denote the number of tokens and the embedding dimension, respectively. Note, the dimensions of fully connected layer inside TB is chosen such that it introduces sufficient data variability. Data variability ensures that the TReX-optimized ViT achieves iso-accuracy with the \textit{BaseViT} model. Apart from this, no other architectural modification is performed on the \textit{BaseViT}. 




\subsection{Attention Reuse-based Optimization}
\label{sec:trex_method}
\begin{algorithm}
\caption{Finding Optimal $N_{reuse}$ to meet Target Delay}
\label{alg:step1}
\begin{algorithmic}[1]
\State Initialize ViT and IMC Parameters ($\mathcal{P}$) in Table \ref{tab:trex_params} 
\State $N_{reuse} \gets 0$      \Comment{Set $N_{reuse}$ to 0}
\State $r \gets 0$
\State $\text{delay} \gets \text{TReXSim}(N_{\text{reuse}}=r, \mathcal{P})$
\While{$\text{delay} > \text{target\_delay}$}
    \State $\text{delay} \gets \text{TReXSim}(N_{\text{reuse}}=r, \mathcal{P})$
    \State $r \gets r + 1$
\EndWhile
\State \text{Optimal} $N_{reuse} \gets r$
\end{algorithmic}
\end{algorithm}
\begin{figure*}[h!]
    \centering
    \includegraphics[width=\textwidth]{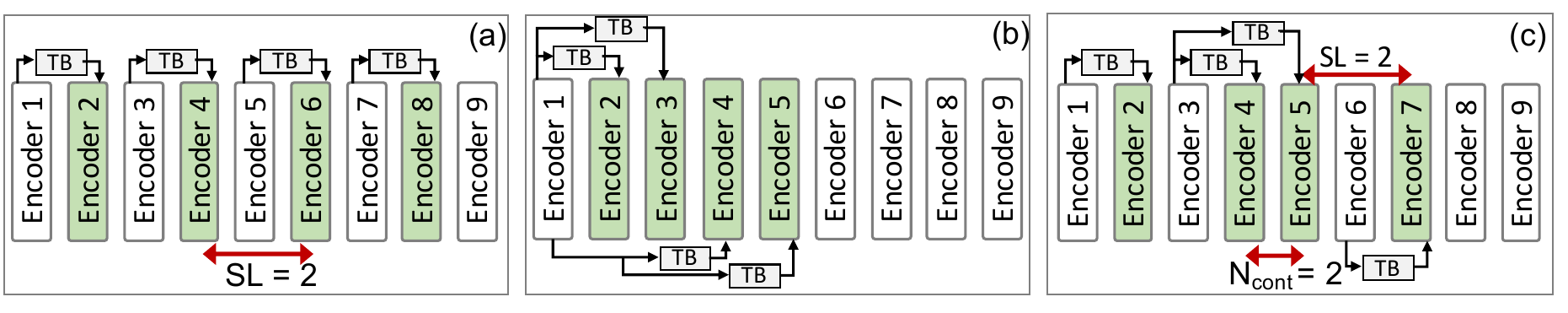}
    \caption{(a) Strided pattern with $SL=2$, $N_{reuse}=4$ (b) Continuous pattern $N_{reuse}=4$ (c) Pyramid pattern with $SL=2$, $N_{reuse}=4$ and $N_{cont}=2$.}
    \label{fig:sharing_patterns}
\end{figure*}
\begin{figure*}
    \centering
    \includegraphics[width=\textwidth]{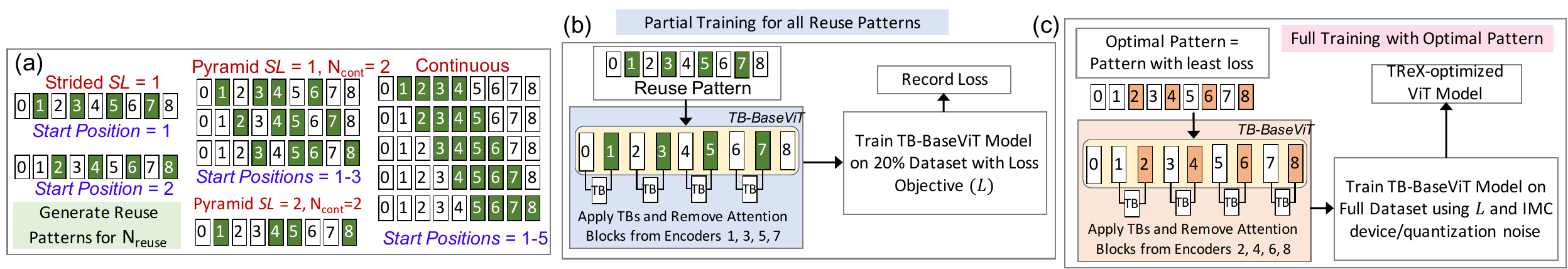}
    \caption{(a) Figure showing an example of different reuse patterns for $N_{Encoders}=9$ and Optimal $N_{reuse}=4$ generated in Step 2. (b) Illustration of Step 3. Here, partial training is performed for all reuse patterns generated in Step 2 (c) Illustration of Step 4. Here, IMC variation-aware training is performed with the optimal pattern.}
    \label{fig:TReX_framework}
\end{figure*}
The goal of the attention reuse-based optimization is to choose optimal number of encoders (Optimal $N_{reuse}$) and their locations in the \textit{BaseViT} model such that high accuracy is achieved while meeting the target delay. This optimization is performed in 4 steps:

\textbf{Step 1: } In this step, we choose the Optimal $N_{reuse}$ value that meets the given target delay criterion (Algorithm \ref{alg:step1}). Before beginning the iteration, first the user-defined ViT and IMC parameters $\mathcal{P}$ shown in Table \ref{tab:trex_params} are initialized. Then, the variable $r$ is iteratively incremented. At each iteration, delay value is computed by the TReX-Sim platform based on the $N_{reuse}$ and $\mathcal{P}$ values. Note, only the $N_{reuse}$ value changes with each iteration while the $\mathcal{P}$ remain constant. $\mathcal{P}$ includes $N_{Encoders}$, $d$, $t$ values among others to compute the delay, $D_{ViT}$, using Equation in row 9 of Table \ref{tab:eda_equations}. The minimum value of $r$ that meets the target delay criteria is chosen and set as the Optimal $N_{reuse}$ value. Note, as the attention block has the major contribution towards the overall delay (Fig. \ref{fig:EDA_Pie_FeFET}b), we choose to optimize the \textit{BaseViT} for a given delay constraint. 

\textbf{Step 2: } While Optimal $N_{reuse}$ affects the energy-efficiency, the reuse pattern (\textit{i.e.,} the location of encoders reusing attention) determines the accuracy of the TReX-optimized ViT as not all encoders are favorable for attention reuse. Since in this work, we use isotropic ViTs such as DeiT-S \cite{rao2021dynamicvit} and LV-ViT-S \cite{yang2022full}, wherein all the encoders have the same architecture, the reuse pattern does not affect the energy, delay or area of the IMC-implemented ViT. 

Exploring all possible reuse patterns for a given Optimal $N_{reuse}$ value leads to a large design space exploration. For example, for a \textit{BaseViT}= DeiT-S with $N_{Encoders}=12$, if Optimal $N_{reuse}=5$, then we need to explore ${11 \choose 5} = 462$ possible reuse patterns. Hence, to reduce the design space, we use well-defined uniform reuse patterns shown in Fig. \ref{fig:sharing_patterns}. 
\begin{enumerate}
    \item \textbf{Strided Pattern: } Shown in Fig. \ref{fig:sharing_patterns}a, the strided pattern involves attention reuse in encoders separated by a uniform distance called the stride length ($SL$). Here, the attention output from the previous encoder is reused in the green encoders with the help of TBs. 
    \item \textbf{Continuous Pattern: }Here, consecutive encoders share the attention from a prior encoder as shown in Fig. \ref{fig:sharing_patterns}b.
    
    \item \textbf{Pyramid Pattern: } The pyramid pattern shown in Fig. \ref{fig:sharing_patterns}c entails a strided-continuous-strided pattern with $N_{cont}$ number of continuous encoders and strided encoders with stride length $SL$.
    
\end{enumerate}

For all the above patterns, the starting position determines the location of the first encoder reusing attention. The strided patterns are generated by varying $SL$ and starting positions. Similarly, different pyramid patterns can be generated by varying $N_{Cont}$, $SL$ and the starting positions. Finally, different continuous patterns can be generated by varying only the starting positions. As an example, Fig. \ref{fig:TReX_framework}a shows all possible patterns generated for a \textit{BaseViT} with $N_{encoders}=9$ and Optimal $N_{reuse}=4$.  

\textbf{Step 3: } Next, partial training for each reuse pattern generated in Step 2 is performed as shown in Fig. \ref{fig:TReX_framework}b. First, a random reuse pattern is selected and applied to the \textit{BaseViT} creating the \textit{TB-BaseViT} model. This is done by adding TBs between suitable encoders and removing the attention blocks in the encoders reusing attention. Then, the \textit{TB-BaseViT} model with attention reuse is trained on 20\% of the Imagenet-1k dataset with loss objective $L$. For \textit{BaseViT}= DeiT-S, we use the loss objective function shown in Equation \ref{eq:loss_deit}. Here, $L_{CE}$ is the cross-entropy loss computed between the predicted and ground-truth labels. $L_{distill}$ is the cross-entropy loss between the predicted labels and the soft-labels from a pre-trained teacher ViT model. $L_{KL}$ is the KL-Divergence loss between the predicted class tokens and class tokens of the pre-trained teacher ViT \cite{touvron2021training}.
Similarly, for \textit{BaseViT}= LV-ViT-S \cite{jiang2021all}, we use the loss objective functions shown in Equation \ref{eq:loss_lvvit}. Here, $L_{dense}$ is the dense score map-based cross-entropy loss \cite{jiang2021all}. $\lambda_1$, $\lambda_2$ and $\lambda_3$ are hyperparameters. For each reuse pattern, the final loss value upon partial training on 20\% dataset is recorded. 
\begin{equation}
    L = L_{CE} + \lambda_1 L_{distill} + \lambda_2 L_{KL}.
    \label{eq:loss_deit}
\end{equation}
\begin{equation}
    L = L_{CE} + \lambda_3 L_{dense}.
    \label{eq:loss_lvvit}
\end{equation}

\textbf{Step 4: }In the final step as shown in Fig. \ref{fig:TReX_framework}c, the reuse pattern with minimum loss value  is selected as the optimal pattern. This optimal pattern is applied to the \textit{BaseViT} model and the \textit{TB-BaseViT} model is trained on the full dataset using the loss objectives shown in Equation \ref{eq:loss_deit} and \ref{eq:loss_lvvit} for DeiT-S and LV-ViT-S \textit{BaseViT}, respectively. 

The training is performed in an IMC-variation aware manner. For a FeFET-based IMC implementation, ADC quantization, read and write noise is introduced in $Q$, $K$, $V$, $Matmul ~(QK^T)$, $Matmul ~\mathcal{S}(QK^T)V$, \textit{Projection} and \textit{MLP} layers during the forward propagation \cite{sun2019impact, agarwal2016resistive}. In contrast, for SRAM-based IMC implementations, only ADC quantization noise is added during the forward propagation as SRAMs are relatively immune to device non-idealities \cite{chen2018neurosim, peng2020dnn+}. The backward propagation, is performed with 16-bit floating point precision. At the end of Step 4, the TReX-optimized ViT model is obtained.

\begin{figure}[h!]
    \centering
    \includegraphics[width=\linewidth]{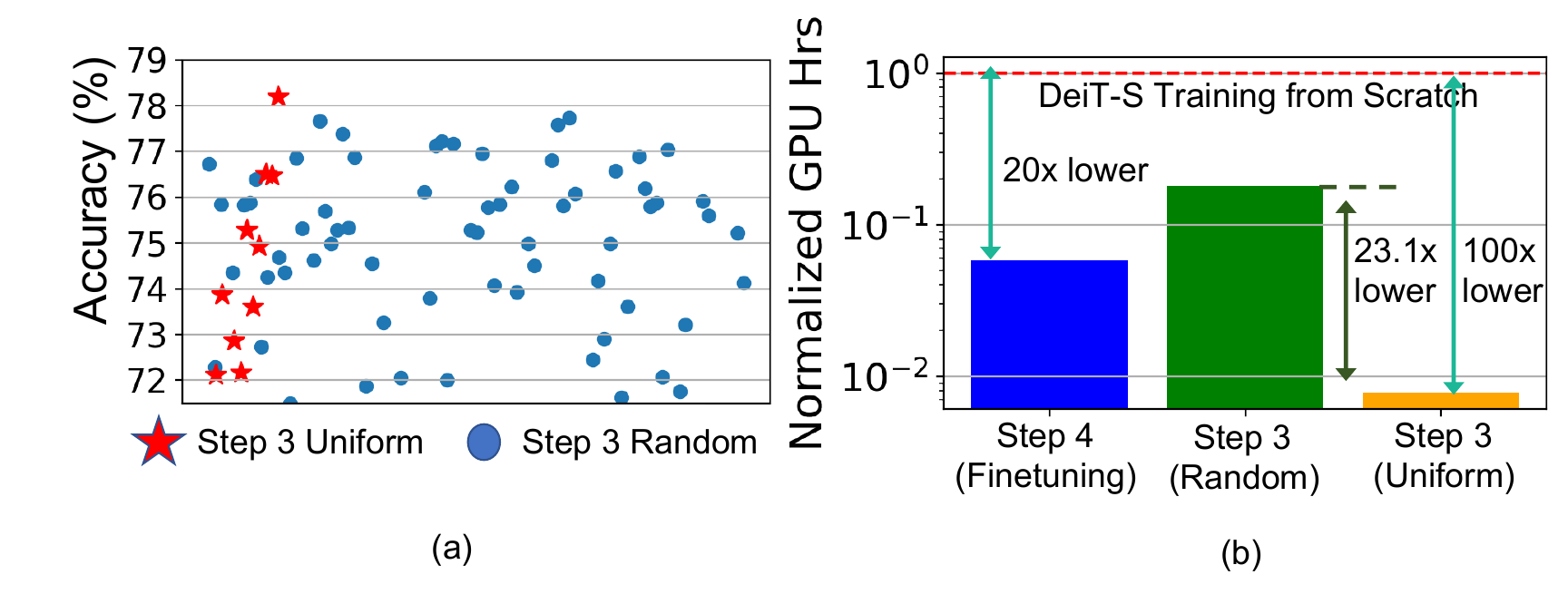}
    \caption{Figure showing (a) Post-finetuning accuracy of DeiT-S ViT with different random and uniform reuse patterns for $N_{reuse}$= 5. The accuracy is measured for ImageNet-1K dataset. (b) Normalized GPU hours (normalized with respect to the cost of training a DeiT-S model from scratch) required by Step 3 ( with random search and uniform pattern search) and Step 4 (finetuning ViT with the optimal reuse pattern.)}
    \label{fig:gpu_cost}
\end{figure}

As shown in Fig. \ref{fig:gpu_cost}a, the uniform reuse patterns achieve higher accuracy compared to different randomly searched patterns for $N_{reuse}$ = 5 for the DeiT-S ViT. This is because the uniform patterns ensure sufficient distance between two encoders reusing the previous encoder's attention. However, this is not always guaranteed in the random reuse patterns. Additionally, since the search space is significantly reduced when uniform reuse pattern is employed, the GPU hours required in the Step 3 of TReX optimization greatly reduces by $ 23.1\times$ compared to the random pattern searching. Note, that Step 3 entails training ViT with different possible reuse patterns over 20\% of the Imagenet-1k dataset. In comparison to training a DeiT-S ViT from scratch, uniform reuse pattern requires 100$\times$ lower GPU hours. With random reuse pattern, the cost is significantly high. Finally, Step 4 (finetuning the ViT with the Optimal reuse pattern) requires 20$\times$ lower GPU hours compared to from-scratch training of DeiT-S.



\section{TReX-Sim Platform}
\label{sec:Trex_sim}
\subsection{Platform Architecture}

\begin{table}[]
\centering
\caption{Table describing the ViT and Crossbar parameters initialized at runtime of the TReX-Sim platform.}
\label{tab:trex_params}
\resizebox{0.8\linewidth}{!}{%
\begin{tabular}{|c|c|}
\hline
\multicolumn{2}{|c|}{ViT Parameters} \\ \hline
$d$                 & Embedding dimension                    \\ \hline
$t$                 & \#Tokens in Input                     \\ \hline
$mlp\_ratio$        & MLP Ratio of ViT                     \\ \hline
$N_{reuse}$          & \#Encoders reusing Attention                     \\ \hline
$N_{Encoders}$        & Total Encoders in ViT                     \\ \hline
$N\_H$              &  Number of Heads in MHSA                    \\ \hline
\multicolumn{2}{|c|}{Crossbar Parameters} \\ \hline
Device              & Type of IMC Device                     \\ \hline
$Xbar\_size$        & Crossbar Size               \\ \hline
$N\_X\_PE$          & \#Crossbars per PE                    \\ \hline
$N\_PE\_Tile$       & \#PE per Tile                     \\ \hline
$E_{R,X}$, $E_{W,X}$ & Crossbar Read and Write Energy  \\ \hline
$D_{R,X}$, $D_{W,X}$ & Crossbar Read and Write Delay  \\ \hline
$A_{X}$ & Crossbar Area \\ \hline
\end{tabular}%
}
\end{table}

\begin{figure*}[h!]
    \centering
    \includegraphics[width=0.9\textwidth]{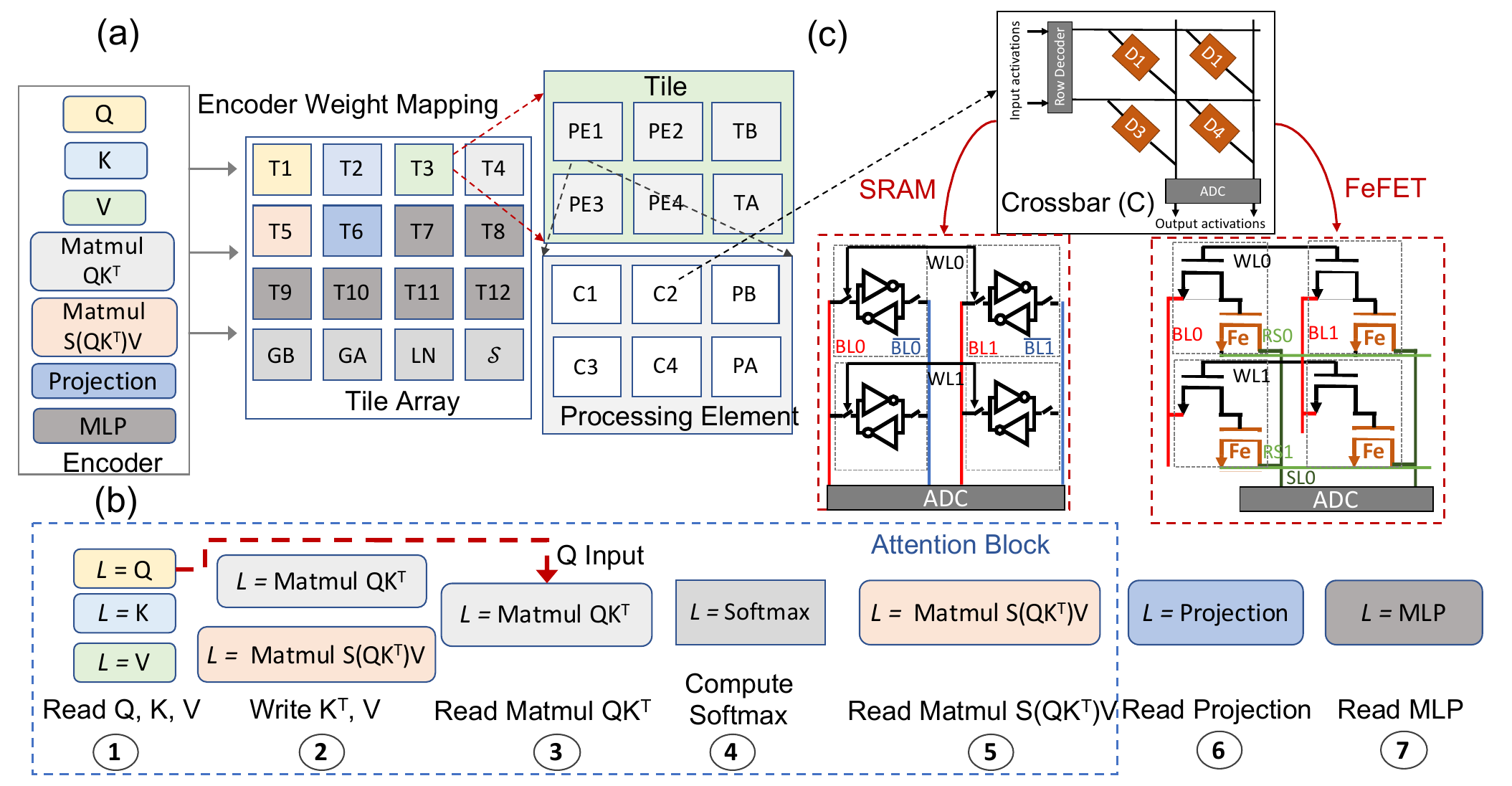}
    \caption{(a) Mapping of encoder layers to Tiles. GB, TB and PB- Global, Tile and PE Buffers. GA, TA, PA- Global, Tile and PE Accumulators. LN-Layer Normalization. (b) The computation stages of an IMC-implemented ViT encoder. (c) SRAM and FeFET Crossbar architectures.}
    \label{fig:arch_fig}
\end{figure*}
\begin{figure}[h!]
    \centering
    \includegraphics[width=0.6\linewidth]{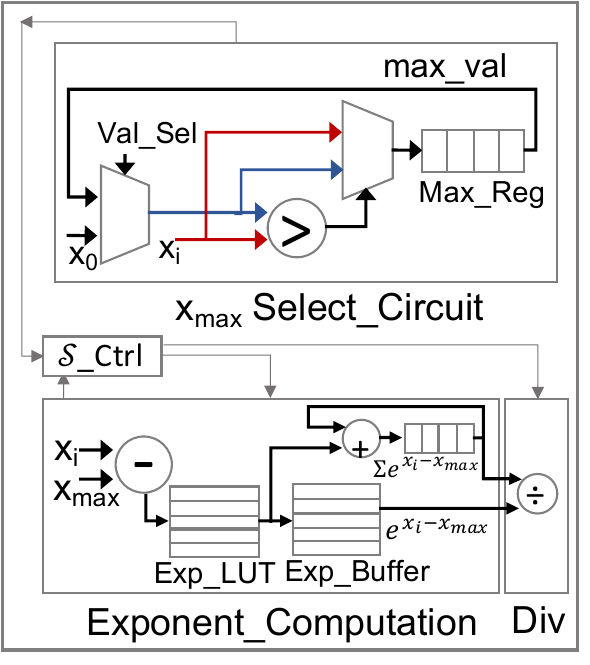}
    \caption{Architecture of the softmax unit. Each softmax module $\mathcal{S}$ contains multiple softmax units dictated by the number of attention heads.}
    \label{fig:softmax}
\end{figure}
TReX-Sim is a hardware-realistic inference benchmarking platform for IMC-implemented ViTs. For evaluation, TReX-Sim is initialized with user-defined ViT and crossbar parameters shown in Table \ref{tab:trex_params}. 
Following this, as shown in Fig. \ref{fig:arch_fig}a, the encoder layers are mapped on a tiled-array architecture. We adopt the standard tiled architecture in prior works \cite{chen2018neurosim, peng2020dnn+}. Following prior works, we assume that multiple tiles can map one layer but not vice-versa \cite{chen2018neurosim, peng2020dnn+}. {Additionally, to reduce the ADC precision, we follow input and weight splitting paradigms similar to prior works \cite{chen2018neurosim, li2020timely}.} The number of crossbars required to map layer $L$ ($L \neq softmax$) is defined by Equation \ref{eq:nxl}. Here, $in\_dim$ and $out\_dim$ are the input and output dimensions of layer $L$, respectively, and crossbar $Xbar\_size$ denotes the crossbar size.

\begin{equation}
    N_{X,~L} = ceil(\frac{in\_dim}{Xbar\_size})\times ceil(\frac{out\_dim}{Xbar\_size})
    \label{eq:nxl}
\end{equation}

Similar to prior IMC-driven transformer works \cite{yang2020retransformer, yang2022full}, the encoder execution occurs as shown in Fig. \ref{fig:arch_fig}b. In \circled{1}, the $Q$, $K$ and $V$ values are read from tiles mapping layers $L=Q$, $L=K$ and $L=V$. In \circled{2}, the $K^T$ and $V$ matrices are written to the tiles mapping $L = Matmul ~QK^T$ and $L = Matmul ~\mathcal{S}(QK^T)V$ layers, respectively. This is followed by crossbar read operations in tiles of $L= Matmul ~QK^T$ to generate the $QK^T$ in \circled{3}. Next, the $\mathcal{S}(QK^T)$ is computed by the digital softmax module ($\mathcal{S}$) in \circled{4}. Following this, read operations are performed in the tiles mapping layers $L = Matmul ~\mathcal{S}(QK^T)V$, $L=Projection$ and $L=MLP$ to compute the $\mathcal{S}(QK^T)V$, projection and the MLP outputs in \circled{5}, \circled{6} and \circled{7}, respectively. Depending on the device, TReX-Sim can support two types of IMC crossbars:
\begin{enumerate}
    \item \textbf{SRAM Crossbar: }The crossbar architecture for the SRAM-based IMC device is shown in Fig. \ref{fig:arch_fig}c (left). The SRAM array is written in a row-by-row manner involving a bit-line pre-charge followed by the write-line (WL) activation to write values into the corresponding SRAM cells. The read is performed in a parallel manner wherein all the WLs are simultaneously activated post the BL pre-charging stage. The analog currents accumulate over the bit lines (BL) and are converted to digital values using flash Analog-to-Digital Converters (ADCs) as shown. 

    \item \textbf{FeFET Crossbar: }The FeFET array has different write and read paths. For the write operation, positive and negative voltage (typically 4V) is supplied to the BL in a series of write pulses. During write, the read-select (RS) line is grounded. For the read operation, read voltage is supplied at the RS line while the BL is grounded. The currents accumulate along the source lines (SL) and are converted to digital values using flash-ADCs. Similar to the SRAM array, the write operation is performed row-by-row while the read is parallel. 
\end{enumerate}



\begin{figure}
    \centering
    \includegraphics[width=0.5\linewidth]{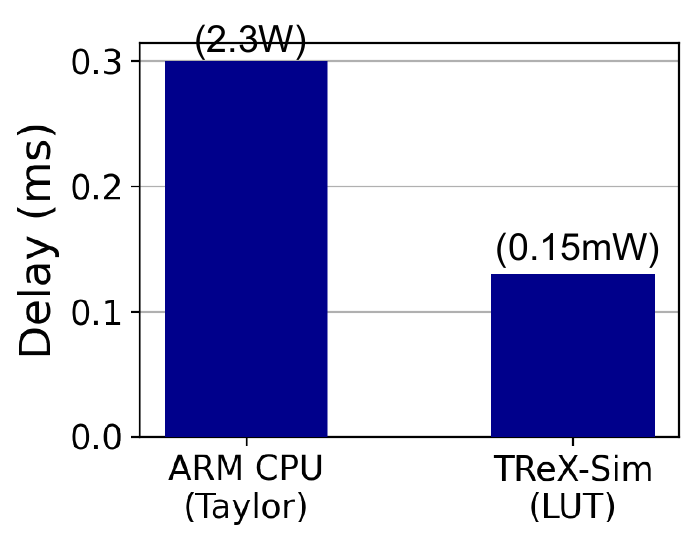}
    \caption{{Figure showing the delay and power of softmax module implemented on ARM CPU (ARM Cortex-A53) and the softmax module integrated into TReX-Sim. The softmax for ARM CPU is implemented using a 6 degree Taylor series approximation (Taylor). Whereas, in TReX-Sim, we use Look-up Tables (LUT) to compute the exponent.}} \vspace{-4mm}
    \label{fig:vfu_compare}
\end{figure}

\textbf{Softmax Module ($\mathcal{S}$): }Fig. \ref{fig:softmax} shows the architecture of the softmax unit. Note, that prior IMC works \cite{yang2022full, yang2020retransformer} have proposed NVM-based softmax modules which require repeated write cycles for softmax computation. However, given the high write energy and device variations in NVMs, NVM-based softmax modules are unreliable. Thus, in TReX, we use a digital softmax module similar to prior works \cite{li2020ftrans, zhang2021algorithm, qi2021accommodating} for reliable and energy-efficient softmax computation. The softmax is computed in a token-by-token basis. For each token, first the $x_{max}$ (refer Equation \ref{eq:softmax}) value is searched. This is performed in the $x_{max}$ \textit{Select\_Circuit} wherein the token features (of dimension $t$) are compared sequentially and the intermediate max values are stored in \textit{Max\_Reg}. Among the $t$ comparison operations, \textit{Val\_Sel} = 1 for the first comparison and 0 for all other comparisons. Following this, the softmax control circuit (\textit{$\mathcal{S}$\_Ctrl}) enables the \textit{Exponent\_Computation} circuit and forwards the $x_{max}$ value. In the \textit{Exponent\_Computation} circuit, $x_{max}$ is subtracted from feature $x_i$ of the token and a lookup is performed on the \textit{Exp\_LUT} to compute the exponent ($e^{x_i-x_{max}}$). Each $e^{x_i-x_{max}}$ is stored in the \textit{Exp\_Buffer} for future computation. The $e^{x_i-x_{max}}$ is also accumulated over all the $t$ tokens. After exponent computation, the \textit{$\mathcal{S}$\_Ctrl} enables the \textit{Div} circuit to compute the softmax values. During this, the exponent values stored in \textit{Exp\_Buffer} are reused. In TReX-Sim, we use independent softmax units for each attention head. For example, for a DeiT-S model with 6 attention heads, the $\mathcal{S}$ module contains 6 softmax units. 

{Due to the Look-up Table (LUT)-based implementation and the parallel computation of softmax across attention heads, TReX-Sim's softmax module is 3$\times$ faster and consumes $10^3\times$ less power compared to CPU-based implementation using Taylor approximation of the exponent function. Notably, the Taylor series approximation is costly due to the large polynomial degree required to approximate the exponential function (polynomial degree = 6). This is shown in Fig. \ref{fig:vfu_compare}.}

\subsection{Hardware Evaluation}
\begin{table}[]
    \caption{Table showing the energy, delay and area equations used by TReX-Sim Platform}
    \label{tab:eda_equations}
    \centering
    
    \resizebox{0.8\linewidth}{!}{
     \begin{tabular}{|l|>{$}c<{$}@{${}{}$}|>{$}c<{$}|}\hline
        1 & E_{R, ~L} & t_L \times N_{X, ~L} \times E_{R, ~X} \\\hline
        2 & E_{W, ~L} & N_{X, ~L} \times E_{W, ~X} \\\hline
        3 & D_{R,L} & t_L \times D_{R,X} \times N_{X,PE} \\\hline
        4 & D_{W,L} & D_{W,X} \times N_{X,PE} \\\hline
        5 & E_{\mathcal{S}} & N_H \times t_L^2 \times (E_{Select}+ E_{Exponent} + E_{Div}) \\\hline
        6 & D_{\mathcal{S}} & t_L^2 \times (D_{Select}+ D_{Exponent} + D_{Div}) \\\hline
        7 & A_{L} & N_{X, ~L} \times A_{X} \\\hline
        8 & E_{ViT} & N_{Encoders}\times(E_{MLP}+ E_{Projection})\\
        & & \quad (N_{Encoders}-N_{reuse})\times E_{Attn} \\ \hline
        
        9 & D_{ViT} & N_{Encoders}\times(D_{MLP}+ D_{Projection})\\
        & & \quad (N_{Encoders}-N_{reuse})\times D_{Attn} \\ \hline
        
        10 & A_{ViT} & N_{Encoders}\times(A_{MLP}+ A_{Projection})\\
        & & \quad (N_{Encoders}-N_{reuse})\times A_{Attn} \\ \hline
    \end{tabular}}
    
\end{table}

Table \ref{tab:eda_equations} shows the equations used by TReX-Sim to compute the read and write energies ($E_{R,~L}$, $E_{W,~L}$), delays ($D_{R,~L}$, $D_{W,~L}$) and area ($A_L$) for any encoder layer $L$ ($L\neq ~Softmax$) shown in Fig. \ref{fig:arch_fig}b. These equations are computed based on the ViT and crossbar parameters shown in Table \ref{tab:trex_params}. As an example, to compute the write energy of $Matmul ~QK^T$ layer, $N_{X,~L}$ is the number of crossbars required to map the $K^T$ vector with $in\_dim$= $t$ and $out\_dim$= $d$. $N_{X,~L}$ multiplied with $E_{W,~X}$ results in $E_{W,L}$ for $Matmul ~QK^T$ layer. 

Additionally, $E_{\mathcal{S}}$ and $D_{\mathcal{S}}$ denote the energy and delay of the softmax module, respectively. Here, $E_{Select}$/$D_{Select}$, $E_{Exponent}$/$D_{Exponent}$ and $E_{Div}$/$D_{Div}$ are the energy/delay of \textit{Select\_Circuit}, \textit{Exponent\_Computation} and \textit{Div} circuits, respectively. Note, the area consumption of the digital softmax module is negligible and hence not considered.
Based on the value of $N_{Encoders}$ and $N_{reuse}$, TReX-Sim computes the ViT energy ($E_{ViT}$), delay ($D_{ViT}$) and area ($A_{ViT}$). Here, $E_{Attn}$, $D_{Attn}$ and $A_{Attn}$ are the summation of read/write energy, delay and area of all layers (including softmax-based computations) in stages 1-5 shown in Fig. \ref{fig:arch_fig}b. 

\section{Results}
\subsection{Experimental Setup}
\textbf{TReX-Optimization Platform: }The baseline and TReX-optimized DeiT-S and LV-ViT-S models are trained using Pytorch 1.6 with Nvidia-V100 GPU backend. We utilize DeiT-S and LV-ViT-S as backbone networks in our approach due to their compact nature and widespread usage in previous studies. During Step 3 of the attention reuse-based optimization, the ViTs are trained on 20\% of the Imagenet-1k dataset. This only adds $\sim$5\% to the overall training overhead. In Step 4 of TReX optimization, after the optimal pattern is applied, the \textit{TB-BaseViT} model is finetuned for 30 epochs on the Imagenet-1k dataset \cite{he2015delving} with a learning rate of 0.01. The hyperparameters $\lambda_1$ and $\lambda_2$ and $\lambda_3$ in Equation \ref{eq:loss_deit} and Equation \ref{eq:loss_lvvit} are 0.1, 0.5 and 0.2, respectively. 

For FeFET-based implementation, read/write device variations shown in Table \ref{tab:xbar_params} and ADC quantization noise are added. For SRAM-based implementations, only ADC quantization noise is added. {While there exists other IMC-specific non-idealities such as IR-drop \cite{moitra2023spikesim} and transistor non-linearities \cite{chakraborty2020geniex}, we use read/write variations (only for FeFET implementation) and ADC quantization noise (for both FeFET and SRAM implementations) to evaluate TReX. Extensive research has shown that device variations and ADC quantization contribute significantly towards accuracy degradation and cannot be easily mitigated \cite{bhattacharjee2023examining, rasch2023hardware}. The IR-drop and transistor non-linearity noise follow a structured profile and can be mitigated using simple approaches such as batchnorm adaptation and weight retraining \cite{bhattacharjee2022examining, bhattacharjee2023examining}}. 



\textbf{TReX-Sim Platform: }Table \ref{tab:xbar_params} shows the hardware parameters used for the energy, delay and area computations of the \textit{BaseViTs} and the TReX-optimized ViTs. The crossbar energy, delay and area values are obtained based on IMC implementations of SRAM and FeFET crossbars of size 64$\times$64 \cite{peng2020dnn+, ni2018memory}. The energy, delay and area of the \textit{Exponent}, \textit{Select} and \textit{Div} circuits in the Softmax unit are computed using 32nm CMOS implementations on the Synopsys DC compiler platform.

\begin{table}[h!]
    \centering
    \caption{Table enlisting the values of various Circuit and Device parameters used for experiments with TReX-Sim.}
    \resizebox{0.7\columnwidth}{!}{
    \begin{tabular}{|l|c|} \hline
       \multicolumn{2}{|c|}{\textbf{Circuit Parameters}} \\ \hline
       Technology & 32nm CMOS  \\ \hline
        $Xbar\_size$, N\_X\_PE, N\_PE\_Tile & 64, 8, 8 \\ \hline
        ADC Precision & 6-bits \\ \hline
        Weight \& Input Precision & 8-bit \& 8-bit\\ \hline
        Input Splitting & 1-bit \\ \hline
        
       \multicolumn{2}{|c|}{\textbf{FeFET Parameters} \cite{ni2018memory}} \\ \hline
       Bits/Cell & 2 \\ \hline
       Read \& Write Variations & 10\% \& 20\% \\ \hline
       $R_{on}$ and $R_{off}$ &  100k$\Omega$ and 10M$\Omega$\\ \hline
       $E_{R,~X}$, $E_{W,~X}$ & 25pJ \& 118pJ\\ \hline
       $D_{R,~X}$, $D_{W,~X}$ & 0.02$\mu$s, 3.3$\mu$s \\ \hline
       $A_X$ & 0.03mm$^2$ \\ \hline
        \multicolumn{2}{|c|}{\textbf{SRAM Parameters} \cite{peng2020dnn+}} \\ \hline
       Bits/Cell & 1 \\ \hline
       $E_{R,~X}$, $E_{W,~X}$ & 29pJ \& 13pJ\\ \hline
       $D_{R,~X}$, $D_{W,~X}$ & 0.018$\mu$s, 0.018$\mu$s \\ \hline
       $A_X$ & 0.07mm$^2$ \\ \hline
    \end{tabular}}
    
    \label{tab:xbar_params}
\end{table}
\begin{table*}[h]
\centering
\caption{Energy, Delay, Area, Accuracy, EDAP, TOPS/W and TOPS/mm$^2$ for DeiT-S and TReX-optimized DeiT-S ViTs optimized at different target delay for the Imagenet-1k dataset. Here, I and NI denote the ideal and non-ideal accuracy (under ADC quantization, read and write variation noise), respectively. TReX-D-$N$ denotes the TReX-optimized DeiT-S Vit at target delay $N$.}
\resizebox{\textwidth}{!}{
\begin{tabular}{|l|c|c|c|c|c|c|c|c|c|}
\hline
Model & Target Delay (ms) & Energy (mJ) & Delay (ms) & Area (mm$^2$) & EDAP & TOPS/W & TOPS/mm$^2$ & \multicolumn{2}{c|}{Accuracy}  \\
\cline{9-10}
& & & & & & & & I & NI \\
\hline
DeiT-S & N/A &  0.13 & 10.92 & 775.2 & 1115.23 (1$\times$) & 34.45 (1$\times$) & 0.00054 (1$\times$) & 79.6 & 79.2 \\
\hline

TReX-D-9 & 9 &   0.12 & 8.46 & 701.1 & 700.2 ($\downarrow$ 1.6$\times$) & 35.49 ($\uparrow$ 1.03$\times$) & 0.00077 ($\uparrow$ 1.42$\times$)& 79.3 & 79 \\ 
\hline

TReX-D-7 &  7 &  0.11 & 6.82 & 651.7 & 484.15 ($\downarrow$ 2.3$\times$) & 36.32 ($\uparrow$ 1.03$\times$)& 0.00102 ($\uparrow$ 1.85$\times$)& 78.4 & 78.1 \\
\hline

TReX-D-6 &  6 &  0.1 & 5.18 & 602.3 & 311.43 ($\downarrow$ 3.5$\times$) & 37.31 ($\uparrow$ 1.05$\times$)& 0.00145 ($\uparrow$ 2.63$\times$)& 77.7 & 77.3 \\
\hline

TReX-D-4 &  4 &  0.091 & 3.54 & 552.9 & 177.6 ($\downarrow$ 6.3$\times$) & 38.49 ($\uparrow$ 1.11$\times$)& 0.00232 ($\uparrow$ 4.21$\times$)& 75 & 74.5 \\
\hline

\end{tabular}}

\label{table:deits-results}
\end{table*}
\begin{table*}[h]
\centering
\caption{Energy, Delay, Area, Accuracy, EDAP, TOPS/W and TOPS/mm$^2$ for LV-ViT-S and TReX-optimized LV-ViT-S ViTs optimized at different target delay for the Imagenet-1k dataset. I and NI denote the ideal accuracy and non-ideal accuracy, respectively. TReX-L-$N$ denotes the TReX-optimized LV-ViT-S Vit at target delay $N$.}
\resizebox{\textwidth}{!}{
\begin{tabular}{|l|c|c|c|c|c|c|c|c|c|}
\hline
Model & Target Delay (ms) & Energy (mJ) & Delay (ms) & Area (mm$^2$) & EDAP & TOPS/W & TOPS/mm$^2$ & \multicolumn{2}{c|}{Accuracy}  \\
\cline{9-10}
& & & & & & & & I & NI \\
\hline
LV-ViT-S & N/A &  0.153 & 14.55 & 912.0 & 2030.1 (1$\times$) & 33.51 (1$\times$) & 0.00039 (1$\times$) & 83.3 & 82.9 \\
\hline

TReX-L-12 &  12 &  0.135 & 11.28 & 813.2 & 1234.87 ($\downarrow$ 1.64$\times$) & 34.59 ($\uparrow$ 1.03$\times$)& 0.00056 ($\uparrow$ 1.43$\times$) & 82.9 & 82.3 \\ 
\hline

TReX-L-10 &  10 &  0.125 & 9.64 & 763.8 & 924.03 ($\downarrow$ 2.19$\times$) & 35.25 ($\uparrow$ 1.05$\times$)& 0.0007 ($\uparrow$ 1.79$\times$) & 82.6 & 82.1 \\
\hline

TReX-L-9 &  9 &  0.121 & 8.82 & 739.1 & 788.35 ($\downarrow$ 2.57$\times$ & 35.61 ($\uparrow$ 1.06$\times$)& 0.00079 ($\uparrow$ 2.02$\times$) & 81.7 & 81.4 \\
\hline

TReX-L-8 &  8 &  0.112 & 7.18 & 689.7 & 553.72 ($\downarrow$ 3.8$\times$) & 36.44 ($\uparrow$ 1.08$\times$)& 0.00103 ($\uparrow$ 2.64$\times$) & 80 & 79.6 \\
\hline

TReX-L-6 &  6 &  0.103 & 5.55 & 640.3 & 364.35 ($\downarrow$ 5.57$\times$) & 37.41 ($\uparrow$ 1.11$\times$)& 0.00144 ($\uparrow$ 3.69$\times$) & 78.1 & 77.9 \\
\hline
\end{tabular}}

\label{table:lvvit-results}
\end{table*}

    


\subsection{Results on DeiT-S and LV-ViT-S Transformers}
As seen in Table \ref{table:deits-results} at near-iso accuracy ($<$2\% NI-accuracy reduction), TReX-D-9, TReX-D-7 and TReX-D-6 achieve significant reduction in the EDAP (upto 3.5$\times$) and increase in the TOPS/W and TOPS/mm$^2$ values by upto 1.05$\times$ and 2.63$\times$, respectively. At a slight NI-accuracy degradation ($\sim$4.7\%) TReX-D-4 achieves 6.3$\times$ EDAP reduction at 1.11$\times$ (4.21$\times$) higher TOPS/W (TOPS/mm$^2$) values compared to DeiT-S. For the LV-ViT-S ViT model (Table V), TReX-L-12, TReX-L-10, TReX-L-9 achieve upto 2.57$\times$ EDAP reduction and upto 1.06$\times$ (2.02$\times$) higher TOPS/W (TOPS/mm$^2$) with a minimal NI-accuracy drop ($\sim$1.5\%) compared to the LV-ViT-S model. With slightly higher accuracy drops (3-5\%), TReX-L-8 and TReX-L-6 achieve upto 5.57$\times$ EDAP reduction and upto 1.11$\times$ (3.69$\times$) higher TOPS/W (TOPS/mm$^2$) compared to LV-ViT-S. Note, we utilize DeiT-S and LV-ViT-S as backbone networks in our approach due to their compact nature and widespread usage in previous studies. 



\begin{figure}
    \centering
    \includegraphics[width=0.8\linewidth]{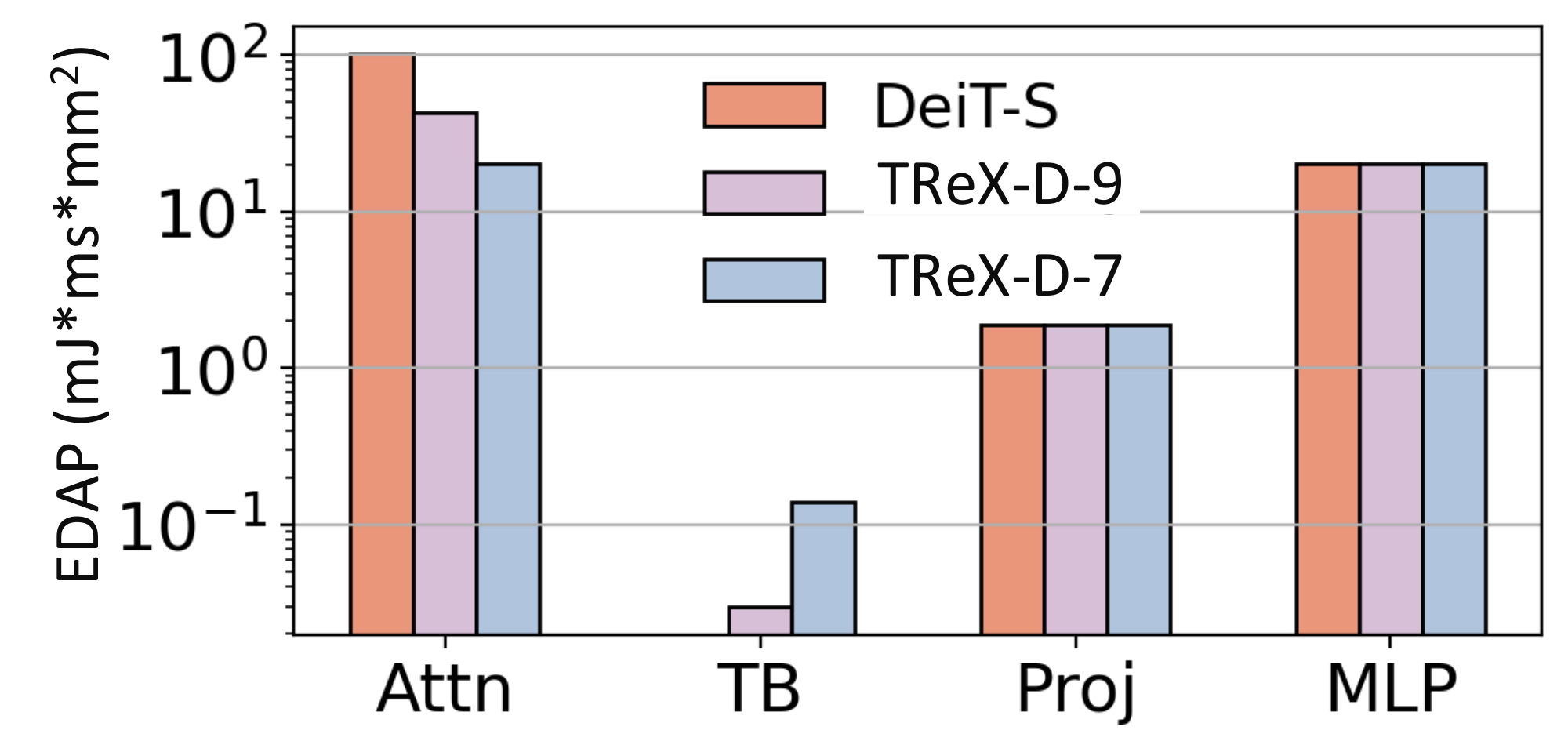}
    \caption{Figure showing the EDAP distribution of DeiT-S, TReX-D-9 and TReX-D-7 models across Attention (Attn), transformation (TB), Projection (Proj) and MLP blocks.}
    \label{fig:edap_distribution}
\end{figure}

\begin{figure}
    \centering
    \includegraphics[width=\linewidth]{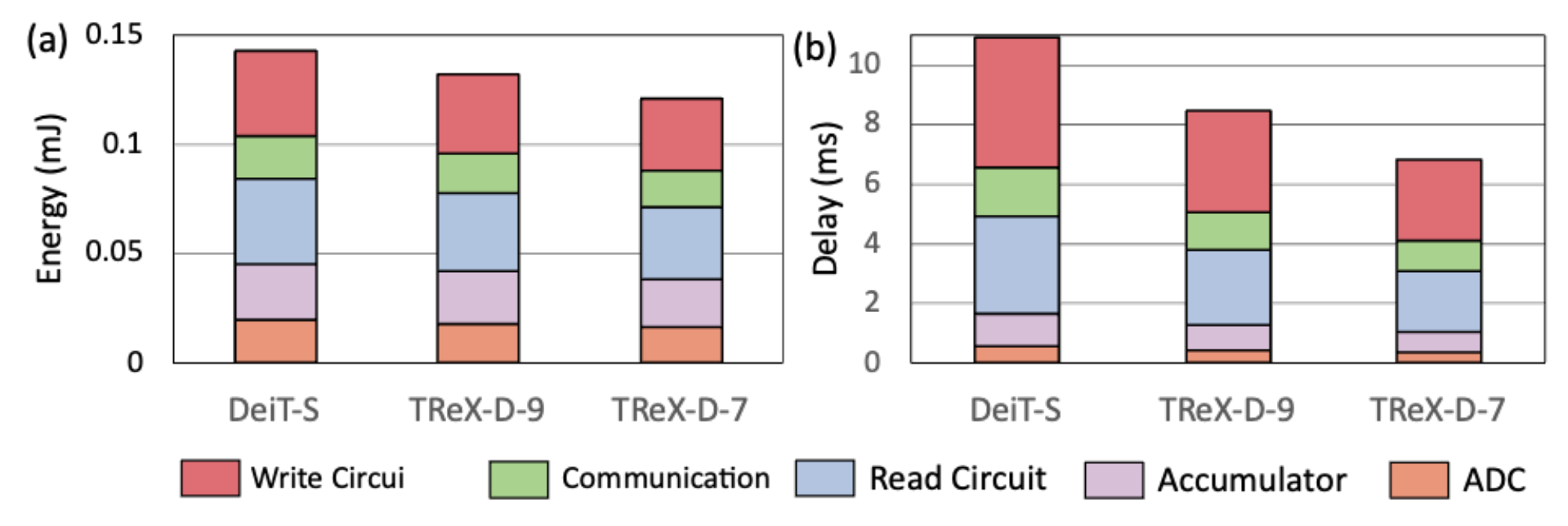}
    \caption{{Figure showing the (a) energy and (b) delay distributions across write circuits, data communication, read circuit, accumulator and ADC. The communication circuit performs intra-tile and inter-tile data communications. While accumulators accumulate partial sums from crossbars, PEs and Tiles.}}
    \label{fig:energy_delay_dist}
\end{figure}

{As the attention module consumes 70-80\% of the overall EDAP (Fig. \ref{fig:EDA_Pie_FeFET}), performing attention reuse majorly reduces the attention overhead which significantly lowers the overall EDAP as seen in Fig. \ref{fig:edap_distribution}.} Although transformation blocks are added for attention reuse, their contribution to EDAP is minimal (less than 1\%). Furthermore, the EDAP of projection and MLP layers remain constant as they are not reused. This further shows that attention reuse has greater capability to minimize the overall EDAP. {From Fig. \ref{fig:energy_delay_dist}, it is observed that TReX optimization lowers the energy and delay across different modules in the TReX-Sim framework.} 

\subsection{Comparison with Prior Works}
\begin{figure}[h!]
    \centering
    \includegraphics[width=0.8\linewidth]{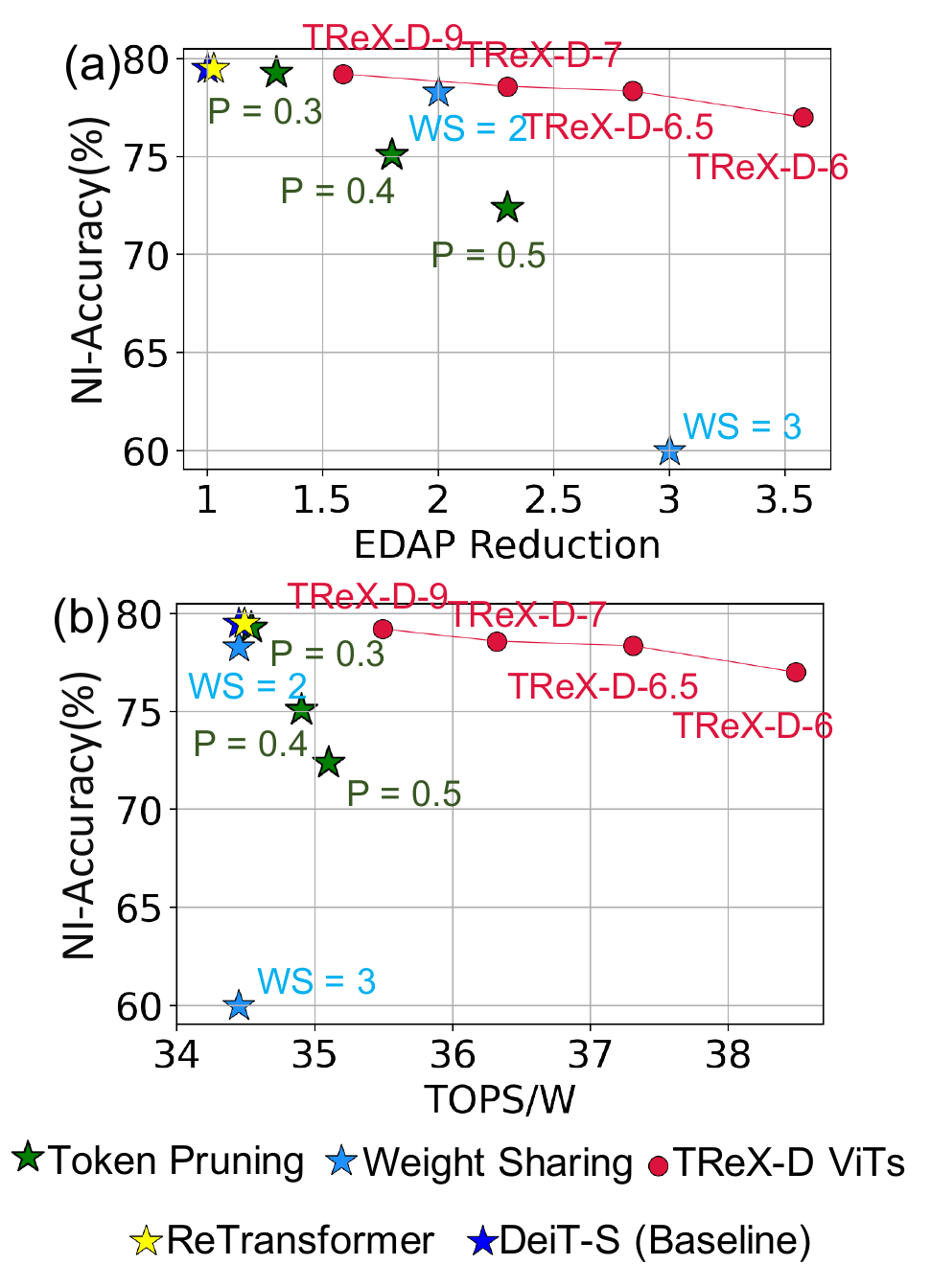}
    \caption{{(a) Figure comparing the non-ideal accuracy and EDAP reduction (relative to the DeiT-S (Baseline) model) between TReX and prior works. (b) Figure comparing the trends in non-ideal accuracy and TOPS/W of TReX and prior works. For token pruning \cite{rao2021dynamicvit}, $P$ represents the token pruning ratio while $WS$ represents the number of repeat encoders for weight sharing \cite{zhang2022minivit}. The TReX ViTs are optimized at 9ms, 7ms, 6.5ms and 6ms. For a fair comparison, all prior optimization works have been implemented on the TReX-Sim framework.} } \vspace{-4mm}
    \label{fig:prior_works}
\end{figure}

Fig. \ref{fig:prior_works}a compares the Accuracy and the EDAP reduction achieved by prior works and TReX. For a fair comparison, all prior works and the baseline DeiT-S model have been implemented on TReX-Sim with 64$\times$64 2-bit FeFET crossbars. ReTransformer \cite{yang2020retransformer} proposed an efficient dataflow to minimize the delay for computing $QK^T$ by eliminating the delay required to program $K^T$. This is done by programming $X^T$ in parallel with $X$-$W_Q$ and $Q$-$W_K^T$ multiplications. Interestingly, as the write delay of FeFET devices is generally $\sim10^3\times$ higher than the read delay \cite{chen2018neurosim}, when ReTransformer is implemented on hardware-realistic IMC-architecture like TReX-Sim, it only reduces the EDAP by 1.03$\times$ compared to baseline while maintaining iso-accuracy. 

Recent works have proposed weight sharing \cite{zhang2022minivit} wherein, multiple encoders in the ViT share the same set of weights (determined by the weight sharing ($WS$) parameter). For example, a $WS$ of 2 implies that 2 encoders share the same weights which means that the number of weight parameters are halved. Weight sharing leads to on-chip area reduction in IMC architectures as less number of crossbars are required to map the weights. However, there is no energy/delay reduction as weight sharing does not reduce the number of computations in the ViT. As shown in Fig. \ref{fig:prior_works}a, for $WS=2$, weight sharing achieves 78.3\% accuracy with 2$\times$ EDAP reduction. Evidently, as $WS$ is increased to 3, the accuracy drastically drops to 60\%.

In token pruning \cite{rao2021dynamicvit}, standalone predictor networks are trained to identify least important tokens that can be pruned out (determined by $P$ parameter) without accuracy loss. Token count reduction directly reduces the number of computations and ultimately the energy/delay. Reduced token count also slightly reduces the number of crossbars required to map $K^T$ and $V$ matrices. However, due to the predictor networks, significant computation overhead is introduced. As seen in Fig. \ref{fig:prior_works}a this limits the EDAP reduction of token pruning to only 1.3$\times$ at 79.3\% accuracy. At higher token pruning ratios ($P>0.3$) token pruning achieves upto 2.2$\times$ EDAP reduction at 7.4\% accuracy drop. Hence, like weight sharing, token pruning is incapable of achieving high EDAP reduction while maintaining high accuracy. 

Finally, TReX is able to beat the state-of-the-art works in terms of EDAP reduction while achieving a high accuracy. TReX-D-9 achieves 1.6$\times$ EDAP reduction at 79.3\% accuracy. Additionally, TReX can maintain high accuracy at high EDAP reduction ($\sim$1.2\% accuracy drop compared to baseline at 2.3$\times$ EDAP reduction). TReX achieves good EDAP-Accuracy tradeoff because it effectively eliminates the redundant attention blocks which as seen in Fig. \ref{fig:EDA_Pie_FeFET}b consumes the major energy-delay-area in a ViT.

{From Fig. \ref{fig:prior_works}b, we observe that TReX achieves better TOPS/W (1.03-1.1$\times$) and accuracy tradeoff compared to prior ViT optimization works. For token pruning \cite{rao2021dynamicvit}, the number of operations and the energy proportionally reduce leading to marginal improvements in the TOPS/W value (1.001-1.01$\times$). When implementing weight sharing \cite{zhang2022minivit} on TReX-Sim, despite the reduction in the number of crossbars, the number of operations and energy consumption remain unchanged. Therefore, the TOPS/W will remain the same as the DeiT-S (Baseline). Finally, ReTransformer entails delay intensive softmax operations leading to a significantly low TOPS/W improvement (1.0001$\times$) compared to DeiT-S (Baseline).}

\begin{figure}[h!]
    \centering
    \includegraphics[width=1\linewidth]{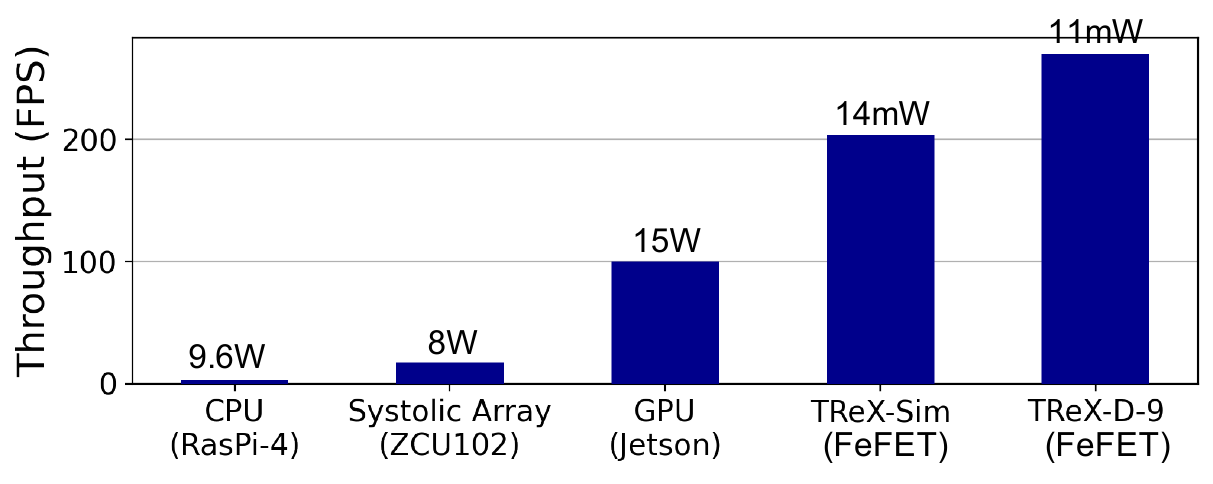}
    \caption{{Plot showing the throughput (measured in Frames-per-second (FPS)) and power consumption of different AI acceleration platforms. FeFET signifies the TReX-Sim and TReX-D-9 implemented with 64x64 FeFET device crossbars. The systolic array accelerator is implemented on a Xilinx ZCU102 FPGA. Raspi-4 and Jetson denote Raspberry-Pi 4 CPU and Nvidia Jetson Orin Nano GPU, respectively.}}
    \label{fig:comp_other_accel}
\end{figure}

{Fig. \ref{fig:comp_other_accel} compares the throughput of a DeiT-S ViT implemented on different low power AI acceleration platforms. Evidently, TReX-Sim (FeFET-based) outperforms CPU, Systolic Array and GPU by 68$\times$, 12$\times$ and 2$\times$, respectively at 10$^3 \times$ lower power. Furthermore, applying the attention reuse optimization (TReX-D-9), improves the throughput by 1.3$\times$ compared to DeiT-S implemented on TReX-Sim (FeFET).}

\subsection{Effect of IMC Device on EDAP-Accuracy Tradeoff}\vspace{-4mm}
\begin{figure}[h!]
    \centering
    \includegraphics[width=\linewidth]{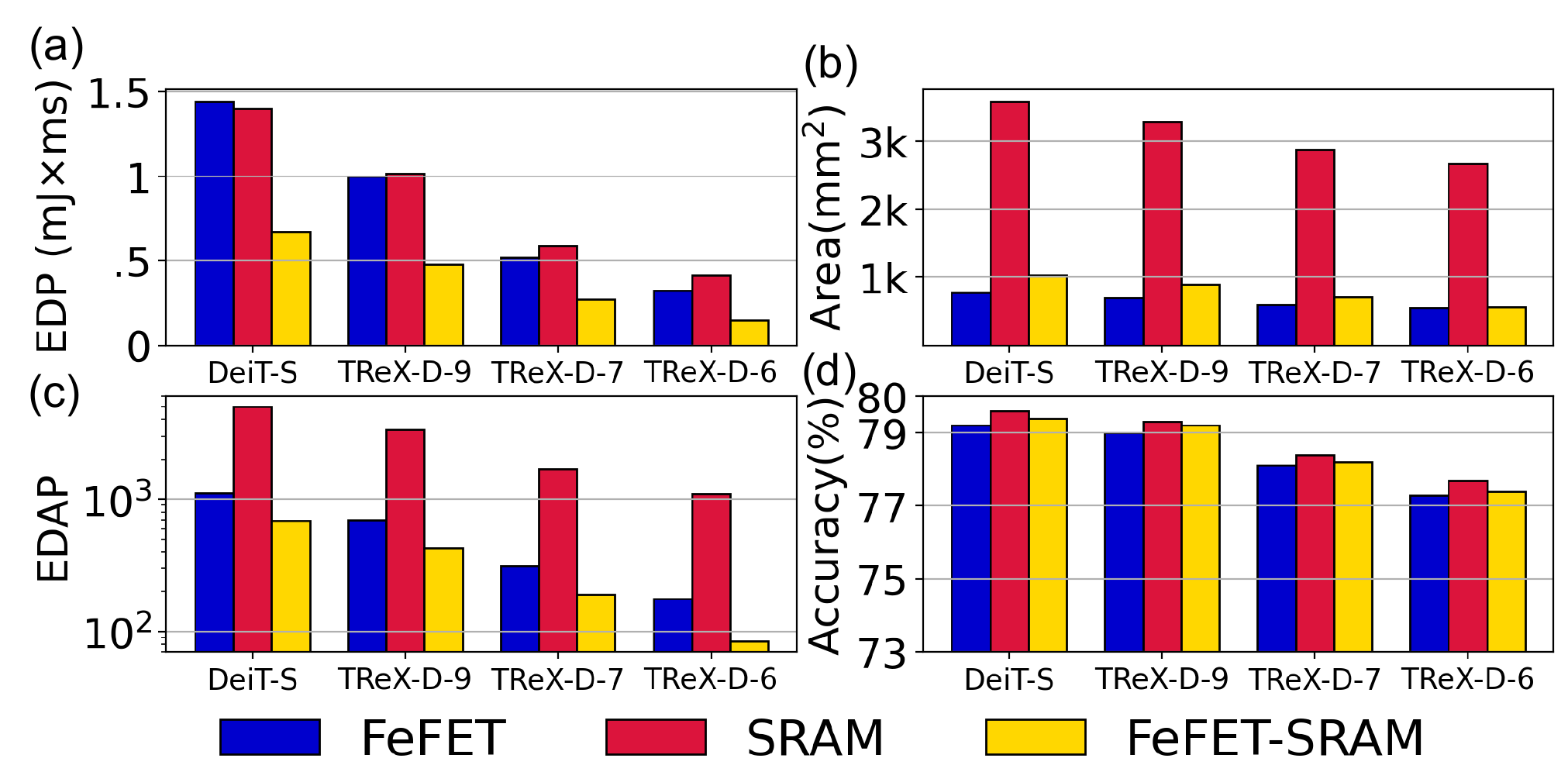}
    \caption{(a) EDP (b) Area (c) EDAP and (d) NI-accuracy of DeiT-S, TReX-D-9 and TReX-D-7 ViT models implemented on 2-bit FeFET, 1-bit SRAM and hybrid FeFET-SRAM IMC architectures. All results correspond to TReX-Sim implementations with 64$\times$64 array.}
    \label{fig:EDAP_Acc_Devices}
\end{figure}
Interestingly, the choice of IMC device plays an important role in the energy-delay-area-accuracy of IMC-implemented ViTs as shown in
Fig. \ref{fig:EDAP_Acc_Devices}. Here, \textit{FeFET} and \textit{SRAM} denote the cases when layers $Q$, $K$, $V$, $Projection$, $MLP$, $Matmul~QK^T$ and $Matmul ~\mathcal{S}(QK^T)V$ are mapped onto 64$\times$64 2-bit FeFET and 1-bit SRAM-based IMC arrays, respectively. 
As seen in Fig. \ref{fig:EDAP_Acc_Devices}a, the EDP of SRAM and FeFET-based implementations are similar. But as FeFET devices can accommodate more bits-per-device, FeFET-based implementations are area efficient compared to SRAM as seen in Fig. \ref{fig:EDAP_Acc_Devices}b leading to an overall lower EDAP (Fig. \ref{fig:EDAP_Acc_Devices}c). However, FeFETs suffer from NVM write and read variations, which leads to low non-ideal accuracy. Typically, the write variations have a more pronounced effect as $K^T$ and $V$ values are periodically written to crossbars mapping layers $Matmul QK^T$ and $Matmul \mathcal{S}(QK^T)V$, respectively. In contrast, SRAM implementations have higher non-ideal accuracy due to the absence of device read or write variations \cite{peng2020dnn+}. 


To achieve the best accuracy-EDAP tradeoff, we propose a FeFET-SRAM hybrid implementation. Here, layers $Matmul ~QK^T$ and $Matmul~\mathcal{S}(QK^T)V$ are mapped onto noise-resilient 1-bit SRAM crossbars while layers $Q$, $K$, $V$, Projection and MLP are mapped onto area-efficient 2-bit FeFET crossbars. This significantly reduces the EDAP while increasing the non-ideal accuracy. Note that layers implemented on FeFET crossbars still suffer from read variations. However, the read variations are much smaller in magnitude compared to write variations \cite{agarwal2016resistive} the non-ideal accuracy evidently improves.  

\subsection{Effect of Reuse Pattern on Non-ideal Accuracy}
\begin{figure}[h!]
    \centering
    \includegraphics[width=\linewidth]{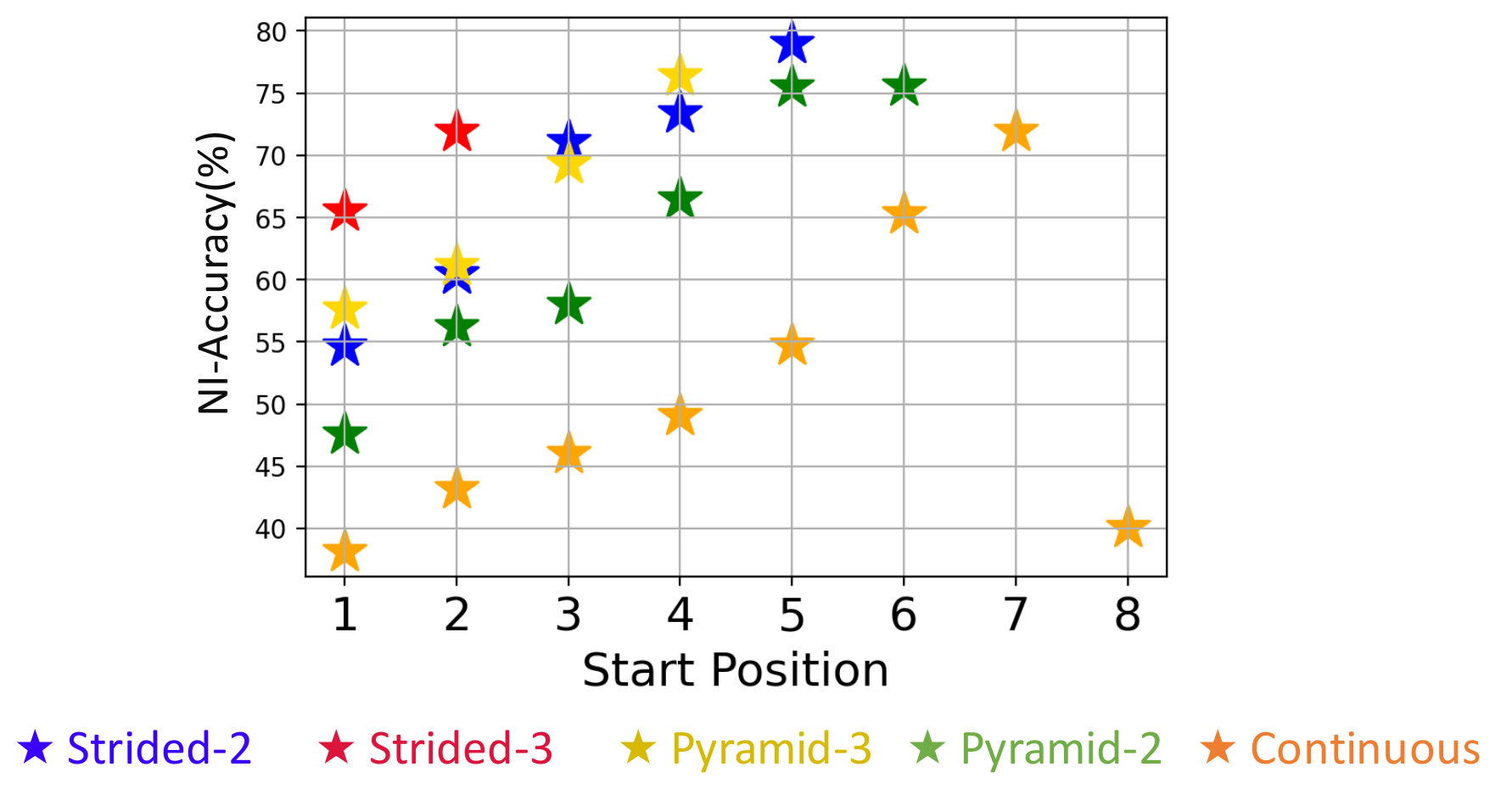}
    \caption{Figure showing the trend of non-ideal accuracy for different reuse patterns applied on \textit{BaseViT}= DeiT-S with $N_{Encoders}=12$ generated and Optimal $N_{reuse}=4$ (for target delay=8ms).}
    \label{fig:pattern_ablation}
\end{figure}

From Fig. \ref{fig:pattern_ablation}, we find that the NI-accuracy depends on the reuse pattern, stride length $SL$ and the start position. It is seen that the accuracy of strided pattern (Strided-2 and Strided-3) is higher than the pyramid. Similarly, the accuracy of pyramid (Pyramid-2 and Pyramid-3) is higher than the continuous pattern. For strided and pyramid patterns, higher $SL$ ensures higher accuracy at the same start position. Interestingly, for the same reuse pattern, higher start position ensures higher accuracy. This suggests that attention reuse in the later encoders are more favorable compared to the attention sharing in the shallow encoders. This is intuitive as the attention in shallow encoders are responsible for extracting closer attention features as compared to the deeper encoders. 

\begin{figure}[h!]
    \centering
    \includegraphics[width=1\linewidth]{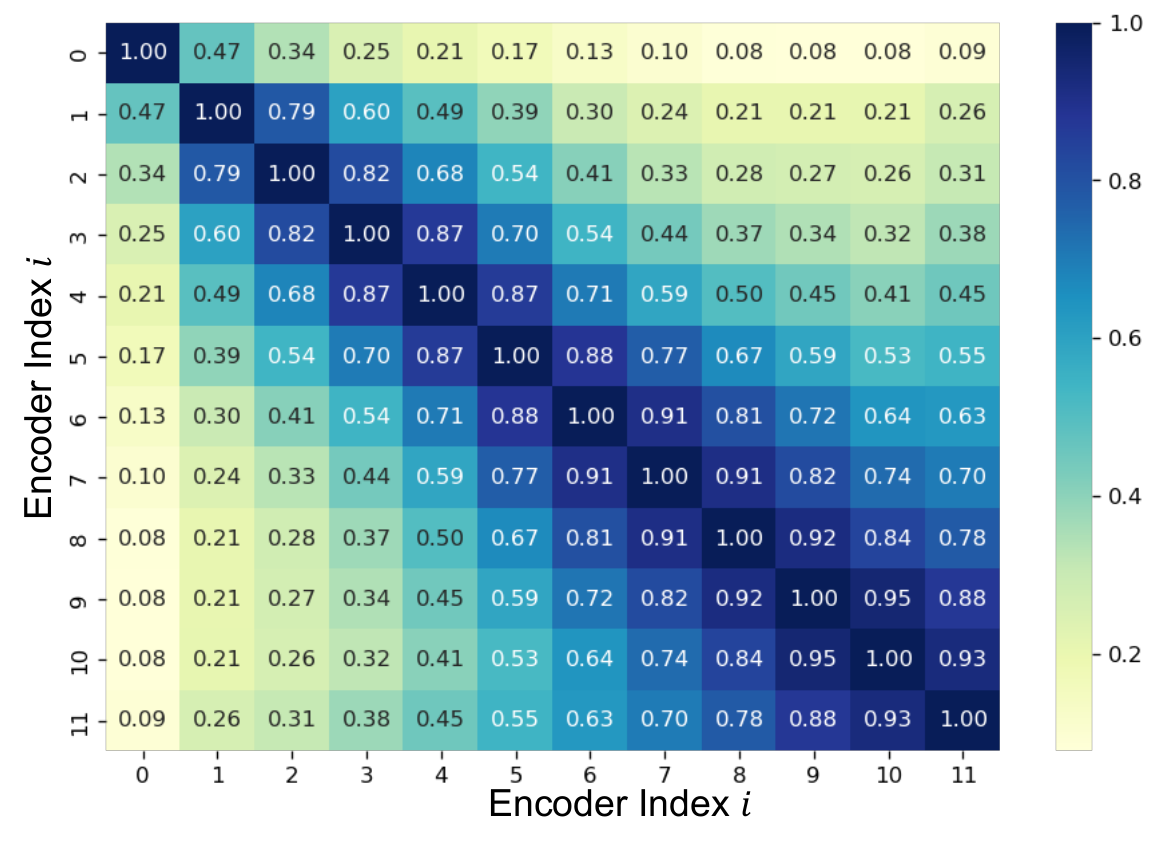}
    \caption{{Center Kernel Alignment (CKA) \cite{cortes2012algorithms} values computed between the attention output of different encoders of the DeiT-S ViT. Each row $i$ corresponds to the CKA value computed between the attention output of the $i_{th}$ encoder with all the other encoders in the ViT. A high CKA value signifies high data correlation and vice-versa. Note, that the CKA of an attention output with itself is always 1.} } \vspace{-5mm}
    \label{fig:cka_heatmap}
\end{figure}

{The variance in accuracy across different reuse patterns can be explained from Fig. \ref{fig:cka_heatmap}. Encoders close together have high correlation between the attention outputs depicted by high Center Kernel Alignment (CKA) values. The correlation reduces as the distance between encoders increases. This suggests that reuse of attentions in a continuous pattern removes important attention information and degrades the accuracy. In contrast, the strided pattern reuses the attention in the adjacent encoder which is highly correlated to the previous one, thereby, leading to better accuracy.}

\subsection{TReX on NLP Task}

\begin{figure}[h!]
    \centering
    \includegraphics[width=1\linewidth]{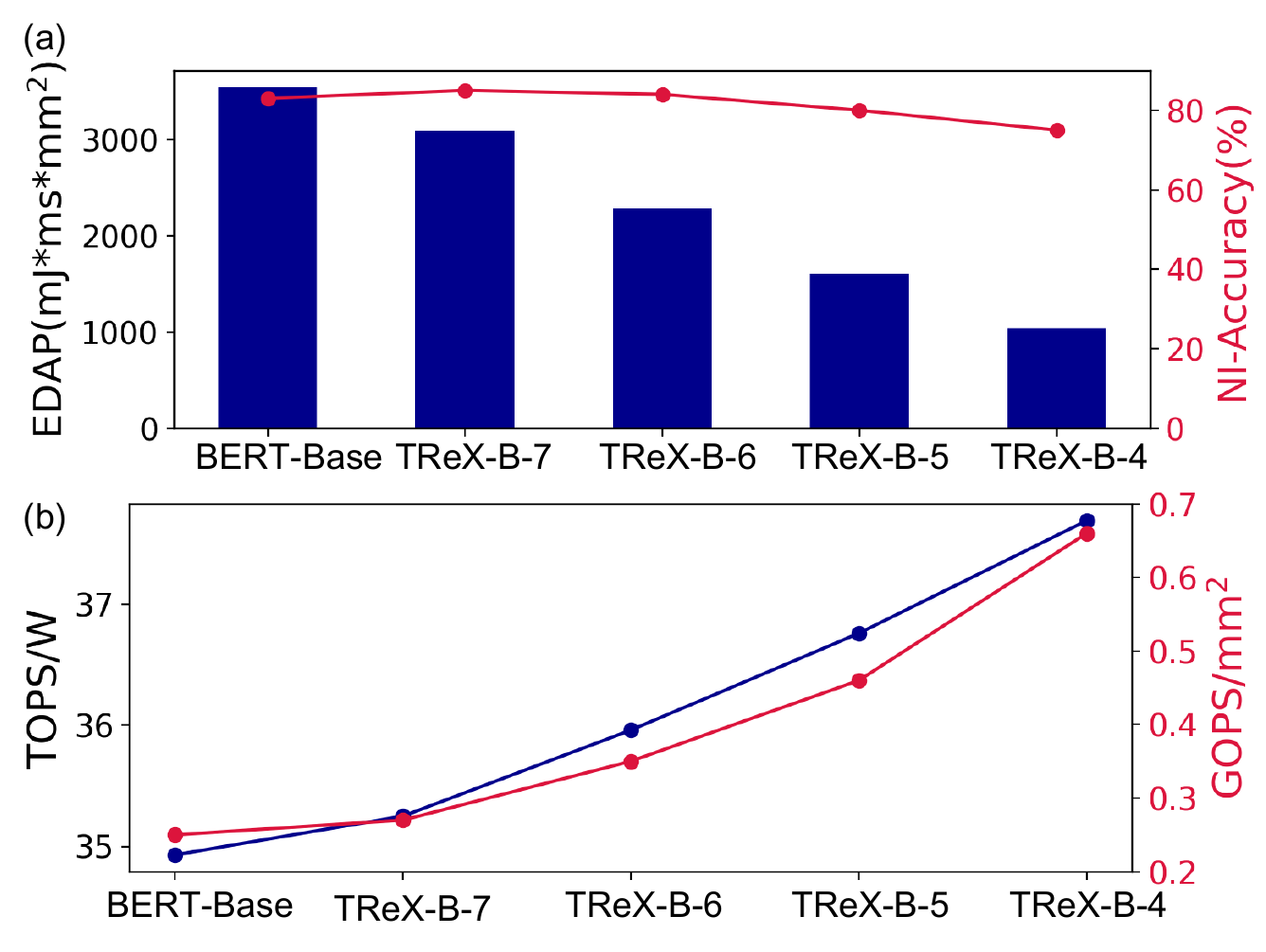}
    \caption{{(a) EDAP and the non-ideal accuracy across different TReX-optimized BERT-Base language model \cite{devlin2018bert} trained on the CoLA \cite{warstadt2018neural} dataset. TReX-B-N denotes TReX-optmized BERT-Base language model at target delay \textit{N}. (b) TOPS/W and TOPS/mm$^2$ values across different TReX-optimized BERT-Base language model. }}
    \label{fig:bert_trex} \vspace{-5mm}
\end{figure}

{To evaluate the attention reuse in TReX on natural language processing (NLP) tasks, we use the CoLA \cite{warstadt2018neural} dataset which consists of 10657 sentences expertly annotated for grammatical acceptability. We apply attention reuse-based optimization (Section \ref{sec:trex_method}) on a pre-trained BERT-Base language model \cite{devlin2018bert} over the CoLA dataset. The TReX-optimized BERT-Base language model is finetuned for 10 epochs after applying attention reuse. As observed in Fig. \ref{fig:bert_trex}a, TReX yields 2$\times$ lower EDAP at 3\% reduction in accuracy in case of TReX-B-5 transformer. In fact, for the TReX-B-7, TReX-B-6 transformer, the accuracy is 2\% higher compared to the BERT-Base language model at an EDAP reduction of 1.6$\times$. At an EDAP reduction of 3.5$\times$, the accuracy drop is 8\% in the TReX-B-4 transformer. TReX also leads to an overall 2.75$\times$ and 1.1$\times$ higher TOPS/mm$^2$ and TOPS/W, respectively (Fig. \ref{fig:bert_trex}b). 
} \vspace{-2mm}

\section{Conclusion}
This work proposes TReX that exploits the unique co-dependence of the ViT attention blocks towards energy-delay-area-accuracy optimization. TReX effectively performs attention sharing in select encoders to achieve high EDAP reduction at minimal loss in non-ideal accuracy compared to \textit{BaseViTs} without attention reuse. TReX-optimized ViTs also achieve higher TOPS/W and TOPS/mm$^2$ compared to the \textit{BaseViTs}. {The efficacy of TReX is demonstrated on both vision and NLP tasks.} In comparison to prior works, TReX-optimized ViTs achieve high EDAP reductions while maintaining high non-ideal accuracy on crossbars. Finally, this work shows the efficacy of FeFET-SRAM hybrid IMC architectures to improve the robustness of IMC-implemented ViT against IMC non-idealities while achieving high EDAP reductions.\vspace{-4mm}

\section*{Acknowledgement}
This work was supported in part by CoCoSys, a JUMP2.0 center sponsored by DARPA and SRC, the National Science Foundation (CAREER Award, Grant \#2312366, Grant \#2318152), and the DoE MMICC center SEA-CROGS (Award \#DE-SC0023198) \vspace{-4mm}

\bibliographystyle{IEEEtran}
\bibliography{TReX.bib}\vspace{-12mm}

\begin{thebibliography}{10}
\providecommand{\url}[1]{#1}
\csname url@samestyle\endcsname
\providecommand{\newblock}{\relax}
\providecommand{\bibinfo}[2]{#2}
\providecommand{\BIBentrySTDinterwordspacing}{\spaceskip=0pt\relax}
\providecommand{\BIBentryALTinterwordstretchfactor}{4}
\providecommand{\BIBentryALTinterwordspacing}{\spaceskip=\fontdimen2\font plus
\BIBentryALTinterwordstretchfactor\fontdimen3\font minus \fontdimen4\font\relax}
\providecommand{\BIBforeignlanguage}[2]{{%
\expandafter\ifx\csname l@#1\endcsname\relax
\typeout{** WARNING: IEEEtran.bst: No hyphenation pattern has been}%
\typeout{** loaded for the language `#1'. Using the pattern for}%
\typeout{** the default language instead.}%
\else
\language=\csname l@#1\endcsname
\fi
#2}}
\providecommand{\BIBdecl}{\relax}
\BIBdecl

\bibitem{touvron2021training}
H.~Touvron, M.~Cord, M.~Douze, F.~Massa, A.~Sablayrolles, and H.~J{\'e}gou, ``Training data-efficient image transformers \& distillation through attention,'' in \emph{International conference on machine learning}.\hskip 1em plus 0.5em minus 0.4em\relax PMLR, 2021, pp. 10\,347--10\,357.

\bibitem{zhai2022scaling}
X.~Zhai, A.~Kolesnikov, N.~Houlsby, and L.~Beyer, ``Scaling vision transformers,'' in \emph{Proceedings of the IEEE/CVF Conference on Computer Vision and Pattern Recognition}, 2022, pp. 12\,104--12\,113.

\bibitem{ranftl2021vision}
R.~Ranftl, A.~Bochkovskiy, and V.~Koltun, ``Vision transformers for dense prediction,'' in \emph{Proceedings of the IEEE/CVF International Conference on Computer Vision}, 2021, pp. 12\,179--12\,188.

\bibitem{maaz2023edgenext}
M.~Maaz, A.~Shaker, H.~Cholakkal, S.~Khan, S.~W. Zamir, R.~M. Anwer, and F.~Shahbaz~Khan, ``Edgenext: efficiently amalgamated cnn-transformer architecture for mobile vision applications,'' in \emph{Computer Vision--ECCV 2022 Workshops: Tel Aviv, Israel, October 23--27, 2022, Proceedings, Part VII}.\hskip 1em plus 0.5em minus 0.4em\relax Springer, 2023, pp. 3--20.

\bibitem{chen2023efficient}
X.~Chen, B.~Kang, D.~Wang, D.~Li, and H.~Lu, ``Efficient visual tracking via hierarchical cross-attention transformer,'' in \emph{Computer Vision--ECCV 2022 Workshops: Tel Aviv, Israel, October 23--27, 2022, Proceedings, Part VIII}.\hskip 1em plus 0.5em minus 0.4em\relax Springer, 2023, pp. 461--477.

\bibitem{moitra2023xpert}
A.~Moitra, A.~Bhattacharjee, Y.~Kim, and P.~Panda, ``Xpert: Peripheral circuit \& neural architecture co-search for area and energy-efficient xbar-based computing,'' \emph{arXiv preprint arXiv:2303.17646}, 2023.

\bibitem{chen2018neurosim}
P.-Y. Chen, X.~Peng, and S.~Yu, ``Neurosim: A circuit-level macro model for benchmarking neuro-inspired architectures in online learning,'' \emph{IEEE Transactions on Computer-Aided Design of Integrated Circuits and Systems}, vol.~37, no.~12, pp. 3067--3080, 2018.

\bibitem{jiang2021all}
Z.-H. Jiang, Q.~Hou, L.~Yuan, D.~Zhou, Y.~Shi, X.~Jin, A.~Wang, and J.~Feng, ``All tokens matter: Token labeling for training better vision transformers,'' \emph{Advances in neural information processing systems}, vol.~34, pp. 18\,590--18\,602, 2021.

\bibitem{rao2021dynamicvit}
Y.~Rao, W.~Zhao, B.~Liu, J.~Lu, J.~Zhou, and C.-J. Hsieh, ``Dynamicvit: Efficient vision transformers with dynamic token sparsification,'' \emph{Advances in neural information processing systems}, vol.~34, pp. 13\,937--13\,949, 2021.

\bibitem{yang2022full}
C.~Yang, X.~Wang, and Z.~Zeng, ``Full-circuit implementation of transformer network based on memristor,'' \emph{IEEE Transactions on Circuits and Systems I: Regular Papers}, vol.~69, no.~4, pp. 1395--1407, 2022.

\bibitem{yang2020retransformer}
X.~Yang, B.~Yan, H.~Li, and Y.~Chen, ``Retransformer: Reram-based processing-in-memory architecture for transformer acceleration,'' in \emph{Proceedings of the 39th International Conference on Computer-Aided Design}, 2020, pp. 1--9.

\bibitem{dong2023heatvit}
P.~Dong, M.~Sun, A.~Lu, Y.~Xie, K.~Liu, Z.~Kong, X.~Meng, Z.~Li, X.~Lin, Z.~Fang, and Y.~Wang, ``Heatvit: Hardware-efficient adaptive token pruning for vision transformers,'' in \emph{2023 IEEE International Symposium on High-Performance Computer Architecture (HPCA)}.\hskip 1em plus 0.5em minus 0.4em\relax IEEE, 2023, pp. 442--455.

\bibitem{zheng2022savit}
C.~Zheng, K.~Zhang, Z.~Yang, W.~Tan, J.~Xiao, Y.~Ren, and S.~Pu, ``Savit: Structure-aware vision transformer pruning via collaborative optimization,'' \emph{Advances in Neural Information Processing Systems}, vol.~35, pp. 9010--9023, 2022.

\bibitem{zhang2022minivit}
J.~Zhang, H.~Peng, K.~Wu, M.~Liu, B.~Xiao, J.~Fu, and L.~Yuan, ``Minivit: Compressing vision transformers with weight multiplexing,'' in \emph{Proceedings of the IEEE/CVF Conference on Computer Vision and Pattern Recognition}, 2022, pp. 12\,145--12\,154.

\bibitem{lin2022spin}
C.-Y. Lin, A.~Prabhu, T.~Merth, S.~Mehta, A.~Ranjan, M.~Horton, and M.~Rastegari, ``Spin: An empirical evaluation on sharing parameters of isotropic networks,'' in \emph{European Conference on Computer Vision}.\hskip 1em plus 0.5em minus 0.4em\relax Springer, 2022, pp. 553--568.

\bibitem{chen2021psvit}
B.~Chen, P.~Li, B.~Li, C.~Li, L.~Bai, C.~Lin, M.~Sun, J.~Yan, and W.~Ouyang, ``Psvit: Better vision transformer via token pooling and attention sharing,'' \emph{arXiv preprint arXiv:2108.03428}, 2021.

\bibitem{zhang2023xformer}
J.~Zhang, Y.~Zhang, J.~Gu, J.~Dong, L.~Kong, and X.~Yang, ``Xformer: Hybrid x-shaped transformer for image denoising,'' \emph{arXiv preprint arXiv:2303.06440}, 2023.

\bibitem{sridharan2023x}
S.~Sridharan, J.~R. Stevens, K.~Roy, and A.~Raghunathan, ``X-former: In-memory acceleration of transformers,'' \emph{IEEE Transactions on Very Large Scale Integration (VLSI) Systems}, 2023.

\bibitem{spoon2021toward}
K.~Spoon, H.~Tsai, A.~Chen, M.~J. Rasch, S.~Ambrogio, C.~Mackin, A.~Fasoli, A.~M. Friz, P.~Narayanan, M.~Stanisavljevic, and G.~W. Burr, ``Toward software-equivalent accuracy on transformer-based deep neural networks with analog memory devices,'' \emph{Frontiers in Computational Neuroscience}, vol.~15, p. 675741, 2021.

\bibitem{xu2022evo}
Y.~Xu, Z.~Zhang, M.~Zhang, K.~Sheng, K.~Li, W.~Dong, L.~Zhang, C.~Xu, and X.~Sun, ``Evo-vit: Slow-fast token evolution for dynamic vision transformer,'' in \emph{Proceedings of the AAAI Conference on Artificial Intelligence}, vol.~36, no.~3, 2022, pp. 2964--2972.

\bibitem{meng2022adavit}
L.~Meng, H.~Li, B.-C. Chen, S.~Lan, Z.~Wu, Y.-G. Jiang, and S.-N. Lim, ``Adavit: Adaptive vision transformers for efficient image recognition,'' in \emph{Proceedings of the IEEE/CVF Conference on Computer Vision and Pattern Recognition}, 2022, pp. 12\,309--12\,318.

\bibitem{shen2022lottery}
X.~Shen, Z.~Kong, M.~Qin, P.~Dong, G.~Yuan, X.~Meng, H.~Tang, X.~Ma, and Y.~Wang, ``The lottery ticket hypothesis for vision transformers,'' \emph{arXiv preprint arXiv:2211.01484}, 2022.

\bibitem{wang2022energy}
Y.~Wang, Y.~Qin, D.~Deng, J.~Wei, Y.~Zhou, Y.~Fan, T.~Chen, H.~Sun, L.~Liu, S.~Wei \emph{et~al.}, ``An energy-efficient transformer processor exploiting dynamic weak relevances in global attention,'' \emph{IEEE Journal of Solid-State Circuits}, vol.~58, no.~1, pp. 227--242, 2022.

\bibitem{wang2021spatten}
H.~Wang, Z.~Zhang, and S.~Han, ``Spatten: Efficient sparse attention architecture with cascade token and head pruning,'' in \emph{2021 IEEE International Symposium on High-Performance Computer Architecture (HPCA)}.\hskip 1em plus 0.5em minus 0.4em\relax IEEE, 2021, pp. 97--110.

\bibitem{stevens2021softermax}
J.~R. Stevens, R.~Venkatesan, S.~Dai, B.~Khailany, and A.~Raghunathan, ``Softermax: Hardware/software co-design of an efficient softmax for transformers,'' in \emph{2021 58th ACM/IEEE Design Automation Conference (DAC)}.\hskip 1em plus 0.5em minus 0.4em\relax IEEE, 2021, pp. 469--474.

\bibitem{moitra2023spikesim}
A.~Moitra, A.~Bhattacharjee, R.~Kuang, G.~Krishnan, Y.~Cao, and P.~Panda, ``Spikesim: An end-to-end compute-in-memory hardware evaluation tool for benchmarking spiking neural networks,'' \emph{IEEE Transactions on Computer-Aided Design of Integrated Circuits and Systems}, 2023.

\bibitem{jain2020rxnn}
S.~Jain, A.~Sengupta, K.~Roy, and A.~Raghunathan, ``Rxnn: A framework for evaluating deep neural networks on resistive crossbars,'' \emph{IEEE Transactions on Computer-Aided Design of Integrated Circuits and Systems}, vol.~40, no.~2, pp. 326--338, 2020.

\bibitem{bhattacharjee2023examining}
A.~Bhattacharjee, A.~Moitra, Y.~Kim, Y.~Venkatesha, and P.~Panda, ``Examining the role and limits of batchnorm optimization to mitigate diverse hardware-noise in in-memory computing,'' in \emph{Proceedings of the Great Lakes Symposium on VLSI 2023}, 2023, pp. 619--624.

\bibitem{chakraborty2020geniex}
I.~Chakraborty, M.~F. Ali, D.~E. Kim, A.~Ankit, and K.~Roy, ``Geniex: A generalized approach to emulating non-ideality in memristive xbars using neural networks,'' in \emph{2020 57th ACM/IEEE Design Automation Conference (DAC)}.\hskip 1em plus 0.5em minus 0.4em\relax IEEE, 2020, pp. 1--6.

\bibitem{bhattacharjee2022examining}
A.~Bhattacharjee, Y.~Kim, A.~Moitra, and P.~Panda, ``Examining the robustness of spiking neural networks on non-ideal memristive crossbars,'' in \emph{Proceedings of the ACM/IEEE International Symposium on Low Power Electronics and Design}, 2022, pp. 1--6.

\bibitem{sun2019impact}
X.~Sun and S.~Yu, ``Impact of non-ideal characteristics of resistive synaptic devices on implementing convolutional neural networks,'' \emph{IEEE JETCAS}, 2019.

\bibitem{agarwal2016resistive}
S.~Agarwal, S.~J. Plimpton, D.~R. Hughart, A.~H. Hsia, I.~Richter, J.~A. Cox, C.~D. James, and M.~J. Marinella, ``Resistive memory device requirements for a neural algorithm accelerator,'' in \emph{2016 International Joint Conference on Neural Networks (IJCNN)}.\hskip 1em plus 0.5em minus 0.4em\relax IEEE, 2016, pp. 929--938.

\bibitem{peng2020dnn+}
X.~Peng, S.~Huang, H.~Jiang, A.~Lu, and S.~Yu, ``Dnn+ neurosim v2. 0: An end-to-end benchmarking framework for compute-in-memory accelerators for on-chip training,'' \emph{IEEE Transactions on Computer-Aided Design of Integrated Circuits and Systems}, vol.~40, no.~11, pp. 2306--2319, 2020.

\bibitem{li2020timely}
W.~Li, P.~Xu, Y.~Zhao, H.~Li, Y.~Xie, and Y.~Lin, ``Timely: Pushing data movements and interfaces in pim accelerators towards local and in time domain,'' in \emph{2020 ACM/IEEE 47th Annual International Symposium on Computer Architecture (ISCA)}.\hskip 1em plus 0.5em minus 0.4em\relax IEEE, 2020, pp. 832--845.

\bibitem{li2020ftrans}
B.~Li, S.~Pandey, H.~Fang, Y.~Lyv, J.~Li, J.~Chen, M.~Xie, L.~Wan, H.~Liu, and C.~Ding, ``Ftrans: energy-efficient acceleration of transformers using fpga,'' in \emph{Proceedings of the ACM/IEEE International Symposium on Low Power Electronics and Design}, 2020, pp. 175--180.

\bibitem{zhang2021algorithm}
X.~Zhang, Y.~Wu, P.~Zhou, X.~Tang, and J.~Hu, ``Algorithm-hardware co-design of attention mechanism on fpga devices,'' \emph{ACM Transactions on Embedded Computing Systems (TECS)}, vol.~20, no.~5s, pp. 1--24, 2021.

\bibitem{qi2021accommodating}
P.~Qi, Y.~Song, H.~Peng, S.~Huang, Q.~Zhuge, and E.~H.-M. Sha, ``Accommodating transformer onto fpga: Coupling the balanced model compression and fpga-implementation optimization,'' in \emph{Proceedings of the 2021 on Great Lakes Symposium on VLSI}, 2021, pp. 163--168.

\bibitem{he2015delving}
K.~He, X.~Zhang, S.~Ren, and J.~Sun, ``Delving deep into rectifiers: Surpassing human-level performance on imagenet classification,'' in \emph{Proceedings of the IEEE international conference on computer vision}, 2015, pp. 1026--1034.

\bibitem{rasch2023hardware}
M.~J. Rasch, C.~Mackin, M.~Le~Gallo, A.~Chen, A.~Fasoli, F.~Odermatt, N.~Li, S.~Nandakumar, P.~Narayanan, H.~Tsai \emph{et~al.}, ``Hardware-aware training for large-scale and diverse deep learning inference workloads using in-memory computing-based accelerators,'' \emph{Nature communications}, vol.~14, no.~1, p. 5282, 2023.

\bibitem{ni2018memory}
K.~Ni, B.~Grisafe, W.~Chakraborty, A.~Saha, S.~Dutta, M.~Jerry, J.~Smith, S.~Gupta, and S.~Datta, ``In-memory computing primitive for sensor data fusion in 28 nm hkmg fefet technology,'' in \emph{2018 IEEE International Electron Devices Meeting (IEDM)}.\hskip 1em plus 0.5em minus 0.4em\relax IEEE, 2018, pp. 16--1.

\bibitem{cortes2012algorithms}
C.~Cortes, M.~Mohri, and A.~Rostamizadeh, ``Algorithms for learning kernels based on centered alignment,'' \emph{The Journal of Machine Learning Research}, vol.~13, pp. 795--828, 2012.

\bibitem{devlin2018bert}
J.~Devlin, M.-W. Chang, K.~Lee, and K.~Toutanova, ``Bert: Pre-training of deep bidirectional transformers for language understanding,'' \emph{arXiv preprint arXiv:1810.04805}, 2018.

\bibitem{warstadt2018neural}
A.~Warstadt, A.~Singh, and S.~R. Bowman, ``Neural network acceptability judgments,'' \emph{arXiv preprint arXiv:1805.12471}, 2018.

\end{thebibliography}

\begin{IEEEbiographynophoto}
{Abhishek Moitra} is pursuing his Ph.D. in the Intelligent Computing Lab at Yale. His research works have been published in reputed journals such as IEEE TCAS-1, IEEE TCAD and conferences such as DAC. His research interests involve hardware-algorithm co-design and co-exploration for designing robust and energy-efficient hardware architectures for deep learning tasks.
\end{IEEEbiographynophoto}\vspace{-13mm}



{\begin{IEEEbiographynophoto}
{Abhiroop Bhattacharjee} is pursuing his Ph.D. in the Intelligent Computing Lab at Yale University. His research interests lie in the area of algorithm-hardware co-design of process in-memory architectures for deep learning \& neuromorphic applications. His research have been published in reputed journals and conferences such as IEEE TCAD, ACM TODAES, DAC, DATE and ISLPED. 
\end{IEEEbiographynophoto}\vspace{-13mm}

{\begin{IEEEbiographynophoto}
{Youngeun Kim} is currently working toward a Ph.D. degree in Electrical Engineering at Yale University, New Haven, CT, USA. Prior to joining Yale,  he worked as a full-time student intern at T-Brain, AI Center, SK Telecom, South Korea. He received his B.S. degree in Electronic Engineering from Sogang University, South Korea, in 2018 and M.S. degree in Electrical Engineering from Korea Advanced Institute of Science and Technology (KAIST), in 2020. His research interests include neuromorphic  computing, computer vision, and deep learning.
\end{IEEEbiographynophoto}\vspace{-13mm}


{\begin{IEEEbiographynophoto}
{Priyadarshini Panda}
obtained her Ph.D. from Purdue University, USA, in 2019. She joined Yale University, USA, as an assistant professor in the Electrical Engineering department in August, 2019. She received the B.E. degree in Electrical and Electronics Engineering  from B.I.T.S. Pilani, India, in 2013. From 2013-14, she worked in Intel, India on RTL design for graphics power management. She has also worked with Intel Labs, USA, in 2017 and Nvidia, India in 2013 as research intern. During her internship at Intel Labs, she developed large scale spiking neural network algorithms for benchmarking the Loihi chip. She is the recipient of the 2019 Amazon Research Award, 2022 Google Research Scholar Award, 2022 DARPA Riser Award, 2022 ISLPED Best Paper Award and 2023 NSF CAREER Award. Her research interests lie in Neuromorphic Computing, Spiking Neural Networks, energy-efficient accelerators, and compute in-memory processing.  
\end{IEEEbiographynophoto}


\end{document}